\documentclass[10pt,twocolumn,letterpaper]{article}

\usepackage[pagenumbers]{cvpr} 

\usepackage{amssymb}

\newcommand{\todo}[1]{{\color{red}#1}}

\usepackage[table,dvipsnames]{xcolor}

\definecolor{ClosedRowBG}{HTML}{EBF0F5}
\definecolor{OpenRowBG}{HTML}{F9FBFD}

\definecolor{ClosedHeadBG}{HTML}{B7C4D6}
\definecolor{OpenHeadBG}{HTML}{D9E5F2}

\definecolor{RankOne}{HTML}{5D87AC}   
\definecolor{RankTwo}{HTML}{80A3C0}   
\definecolor{RankThree}{HTML}{9EBED6} 
\definecolor{RankFour}{HTML}{C0D8E8}  

\newcommand{\ClosedGroup}[1]{%
  \rowcolor{ClosedHeadBG}\multicolumn{#1}{c}{\textit{Proprietary models}}\\
}
\newcommand{\OpenGroup}[1]{%
  \rowcolor{OpenHeadBG}\multicolumn{#1}{c}{\textit{Open-source models}}\\
}
\newcommand{\rankone}[1]{\cellcolor{RankOne}\textbf{#1}}
\newcommand{\ranktwo}[1]{\cellcolor{RankTwo}\underline{#1}}
\newcommand{\rankthree}[1]{\cellcolor{RankThree}#1}
\newcommand{\rankfour}[1]{\cellcolor{RankFour}#1}

\DeclareRobustCommand{\rankswatch}[1]{%
  \raisebox{0.2ex}{\textcolor{#1}{\rule{1.1ex}{1.1ex}}}%
}
\DeclareRobustCommand{\ranklegend}{%
  \rankswatch{RankOne}~
  \rankswatch{RankTwo}~
  \rankswatch{RankThree}~
  \rankswatch{RankFour}%
}

\usepackage{xcolor}
\usepackage{listings}
\usepackage{tcolorbox}
\usepackage[T1]{fontenc}
\usepackage{textcomp}
\tcbuselibrary{skins,breakable,listings}

\definecolor{PromptDeepBlue}{HTML}{0B2F5B}
\definecolor{PromptPaleBlue}{HTML}{F7FAFE}

\newtcolorbox{promptbox}[2][]{
  enhanced,
  breakable,
  sharp corners,
  colback=PromptPaleBlue,
  colframe=PromptDeepBlue,
  colbacktitle=PromptDeepBlue,
  coltitle=white,
  title={\textbf{#2}},
  fonttitle=\bfseries,
  boxrule=0.9pt,
  left=1.5mm,
  right=1.5mm,
  top=1.2mm,
  bottom=1.2mm,
  toptitle=1mm,
  bottomtitle=1mm,
  #1
}

\lstdefinestyle{promptlistingstyle}{
  basicstyle=\ttfamily\scriptsize,
  columns=fullflexible,
  keepspaces=true,
  breaklines=true,
  breakatwhitespace=false,
  showstringspaces=false,
  upquote=true,
  tabsize=2
}

\newtcblisting{promptlisting}[2][]{
  enhanced,
  breakable,
  sharp corners,
  listing only,
  listing options={style=promptlistingstyle},
  colback=PromptPaleBlue,
  colframe=PromptDeepBlue,
  colbacktitle=PromptDeepBlue,
  coltitle=white,
  title={\textbf{#2}},
  title after break={\textbf{#2 (continued)}},
  fonttitle=\bfseries,
  boxrule=0.9pt,
  left=1.5mm,
  right=1.5mm,
  top=1.2mm,
  bottom=1.2mm,
  toptitle=1mm,
  bottomtitle=1mm,
  #1
}

\newcommand{\name}{KeyFrame-Compass\xspace}

\definecolor{cvprblue}{rgb}{0.21,0.49,0.74}
\usepackage[pagebackref,breaklinks,colorlinks,allcolors=cvprblue]{hyperref}
\usepackage{pifont}
\usepackage{makecell}
\usepackage{enumitem}
\usepackage{xcolor}
\usepackage{caption}

\newcommand{\cmark}{\textcolor{green!70!black}{\ding{51}}}
\newcommand{\xmark}{\textcolor{red!80!black}{\ding{55}}}

\def\paperID{*****} 
\def\confName{CVPR}
\def\confYear{2026}

\title{\name: Towards Comprehensive Evaluation of Keyframe-Conditioned Video Generation}

\author{%
    Yuqi Tang$^{1,2}$\,
    Tengfei Liu$^{3}$\,
    Yizheng Lai$^{4}$\,
    Yuran Wang$^{3}$\,
    Yang Shi$^{3,2\ast}$\,
    Wanshun Su$^{5}$\,
    Zhuoran Zhang$^{3}$
    \\
    Qixun Wang$^{3}$\,
    Xiaohan Zhang$^{6}$\,
    Xinlei Yu$^{7}$\,
    Xuehai Bai$^{2}$\,
    Xuanyu Zhu$^{3}$\,
    Bohan Zeng$^{3}$\,
    Bozhou Li$^{3}$
    \\
    Shujie Li$^{8}$\,
    Yifan Dai$^{2}$\,
    Yujie Wei$^{9}$\,
    Shixuan Liu$^{10}$\,
    Haotian Wang$^{10\ast}$\,
    Jialu Chen$^{2}$\,
    Yuanxing Zhang$^{2\ast}$
    \\
    $^{1}$HKUST(GZ) \,
    $^{2}$Kling Team \,
    $^{3}$PKU \,
    $^{4}$RUC \,
    $^{5}$NWPU \,
    $^{6}$NJU \,
    $^{7}$CUHK \,
    $^{8}$HKU \,
    $^{9}$FDU \,
    $^{10}$THU
    \\
    \url{https://github.com/cactusqq/KeyFrame-Compass}
}

\begin{document}

\twocolumn[{
    \renewcommand\twocolumn[1][]{#1}
    \maketitle
    \vspace{-30pt}

    \begin{center}
        \includegraphics[width=\linewidth]{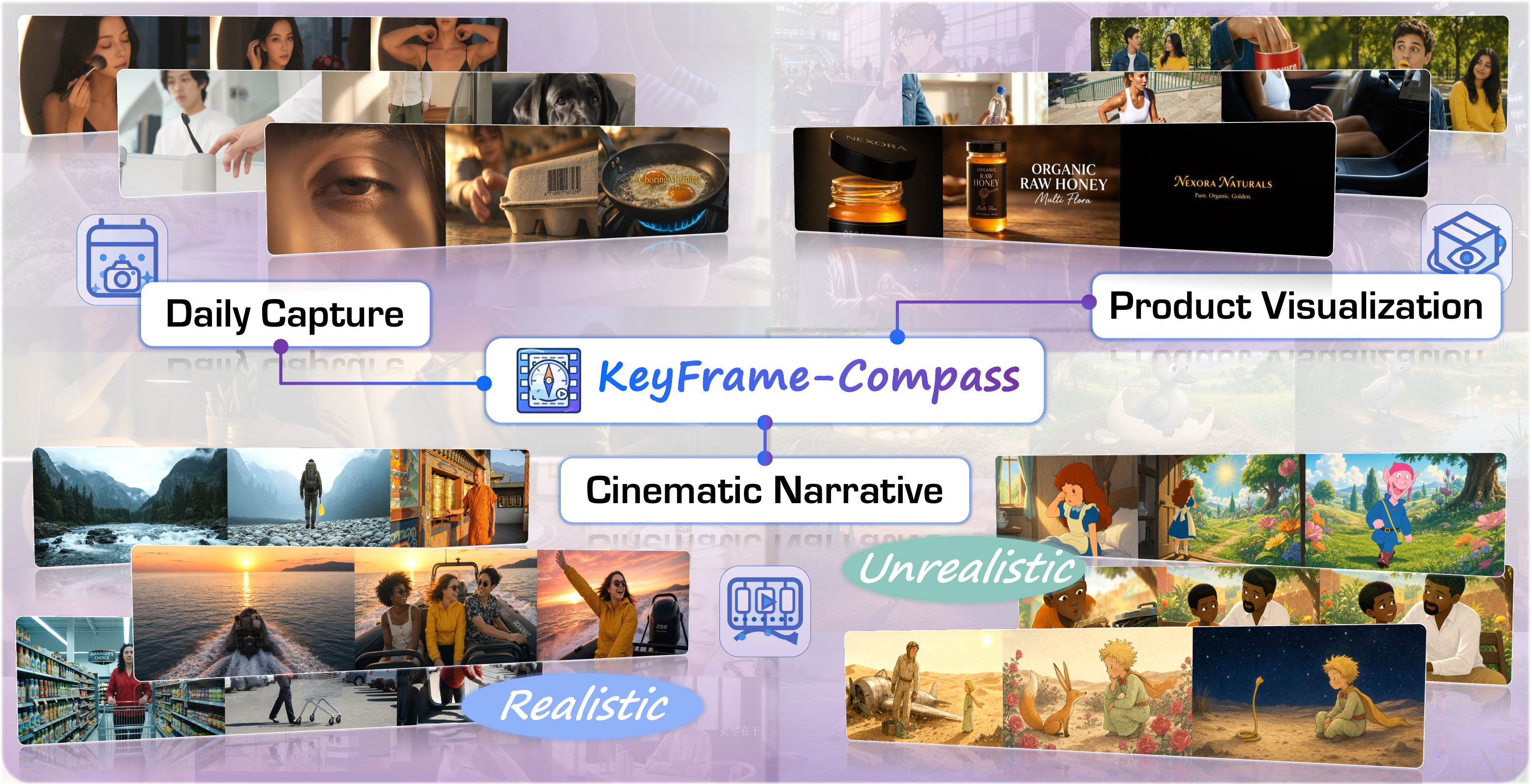}
        \vspace{-12pt}
        \captionof{figure}{
            \textbf{Illustration of \name.}
            The benchmark spans three application domains
            (\textit{daily capture}, \textit{product visualization}, and
            \textit{cinematic narrative}) and covers both realistic and
            stylized visual content. Each test case provides an ordered set
            of keyframes that the generated video must reproduce at the
            intended moments.
        }
        \label{fig:teaser}
        \vspace{2pt}
    \end{center}
}]

\begingroup
\renewcommand{\thefootnote}{\fnsymbol{footnote}}
\footnotetext[1]{Corresponding authors.}
\endgroup

\begin{abstract}
Video generation increasingly relies on keyframe-based workflows, where creators specify a sequence of reference images to guide generation. 
Although recent models support multi-keyframe conditioning, it remains unclear whether they can faithfully reproduce the prescribed keyframes while maintaining overall video quality.
We present \textbf{\name}, the first comprehensive benchmark for evaluating keyframe-conditioned video generation. 
The benchmark contains $386$ carefully curated samples spanning three application domains, two video structures, two prompt granularities, two conditioning formats, and four keyframe densities, enabling controlled analysis under diverse generation settings. 
We further introduce an automated evaluation framework that jointly measures keyframe execution and overall video quality. 
Specifically, we decompose keyframe execution into six complementary metrics covering \textit{presence}, \textit{fidelity}, \textit{temporal ordering}, \textit{localization}, \textit{persistence}, and \textit{uniqueness}, while assessing overall video quality through evidence-grounded MLLM judgments augmented with specialized perception models.
Experiments on nine representative video generation systems reveal several fundamental limitations.
Current models exhibit a clear trade-off between faithful keyframe execution and natural video synthesis. 
Their performance further degrades as keyframe constraints become denser and most open-source models also fail to interpret storyboard-grid inputs as temporally ordered keyframe sequences. 
\end{abstract}

\section{Introduction}
\label{sec:intro}

\begin{figure*}[htbp]
    \centering
    \includegraphics[width=1\linewidth]{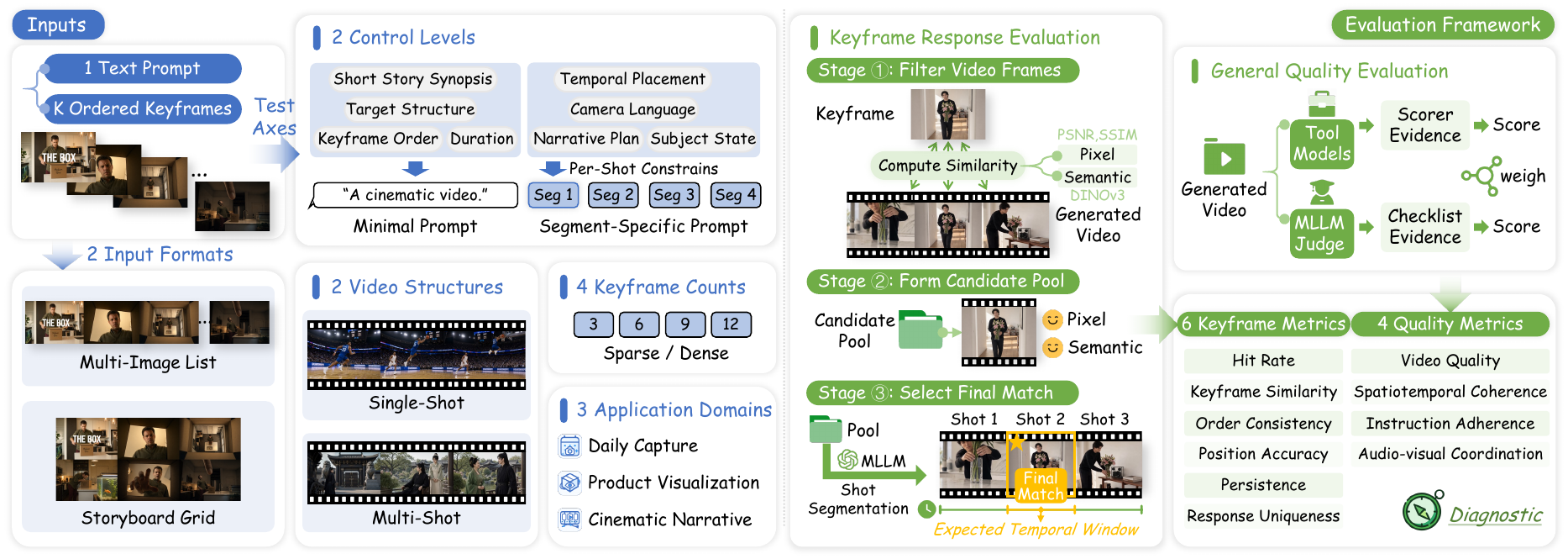}
    \caption{\textbf{Overview of \name.} \textbf{Benchmark Design:} \name evaluates keyframe-conditioned video generation from a text prompt and an ordered set of keyframes, covering diverse domains. \textbf{Evaluation Framework:} Keyframe response is assessed by a staged matching pipeline that combines MLLM-assisted shot segmentation with pixel-level and semantic matching to locate keyframe occurrences within exact temporal windows. General video quality is evaluated by specialized perception models and evidence-grounded MLLM judges. Together, they yield $6$ keyframe-response metrics and $4$ general quality groups for diagnosing controllability, temporal alignment, and overall generation quality.}
    \label{fig:overview}
\end{figure*}

Video generation increasingly relies on keyframe-based workflows, where creators first specify key moments as reference images and then expect the generation model to transform them into a coherent video~\cite{zhang2026stage,xiao2025captain,zhang2026smartdirector}. 
Recent advances in image generation have made creating such visual plans easier than ever, shifting the practical bottleneck from designing keyframes to executing them as complete videos. 
Keyframe-conditioned video generation~\cite{pan2026omniweaving,liu2025dreamontagea} addresses this downstream task by taking an ordered set of reference images together with a text prompt as input and generating a video that faithfully realizes the prescribed visual plan. 
Unlike conventional image-conditioned generation, it must preserve an entire sequence of keyframes, place each one at the correct time, and maintain coherent motion between them.

However, existing benchmarks are not designed to evaluate this setting. General video generation benchmarks mainly assess perceptual quality, temporal consistency, prompt following, and related aspects~\cite{huang2023vbench,sun2024t2v,feng2024tcbenchbenchmarkingtemporalcompositionality,liu2023evalcrafter}. 
Image-conditioned benchmarks instead focus on single-image preservation, spatiotemporal completion, story visualization, and related tasks~\cite{fan2024aigcbench,cai2026videocanvas,zhang2025ui2vbenchunderstandingbasedimagetovideogeneration,zhuang2026vistorybench}.
Although these benchmarks evaluate overall video quality and alignment with textual or visual conditions, none measures whether a complete keyframe sequence is reproduced with the correct appearance, order, and timing.
Nor can this problem be captured by simply adding a single alignment score. 
Multi-keyframe-conditioned generation exhibits several distinct failure modes: keyframes may be omitted, inaccurately reproduced, misplaced in time, presented in the wrong order, or connected by implausible transitions. These errors arise from different underlying abilities, including visual preservation, temporal localization, sequence ordering, and motion synthesis. 
A meaningful benchmark should therefore evaluate each dimension separately rather than collapsing them into a single aggregate score.

To address this gap, we introduce \textbf{\name}, a diagnostic benchmark for multi-keyframe-conditioned video generation. 
Each test case consists of an ordered set of keyframes and a text prompt, requiring the model to reproduce the specified visual sequence while generating coherent transitions between adjacent keyframes. 
As illustrated in Figure~\ref{fig:teaser}, the benchmark uses carefully curated, high-quality keyframes spanning three application domains: daily capture, product visualization, and cinematic narrative. 
It also covers both realistic and stylized visual content.
To systematically examine model performance under varying levels of difficulty, \name contains $386$ evaluation cases spanning five controlled factors, as shown in Figure~\ref{fig:overview} (left): video structure (one-take vs.\ multi-shot), keyframe count ($3$, $6$, $9$, or $12$), prompt specificity (minimal vs.\ segment-specific), conditioning interface (multi-image list vs.\ storyboard grid), and application domain.

Beyond covering diverse task settings, \name evaluates generated videos along two complementary axes: whether the model follows the keyframe plan and whether it produces a high-quality video. 
Accordingly, it reports two groups of metrics, as shown in Figure~\ref{fig:overview} (right). 
The \emph{keyframe response metrics} measure six aspects of plan execution: \textit{presence}, \textit{fidelity}, \textit{ordering}, \textit{timing}, \textit{persistence}, and \textit{uniqueness}. 
The \emph{general quality metrics} combine judgments from multimodal large language models (MLLMs)~\cite{google2026gemini31pro,bai2025qwen3,shi2025mavors} with signals from specialized perception models to assess \textit{video quality}, \textit{spatiotemporal coherence}, \textit{attribute consistency}, \textit{physical plausibility}, \textit{instruction following}, and \textit{audio–video coordination}. 
This separation reveals whether a model fails to execute the prescribed keyframes or to synthesize a coherent and plausible video.

Using this evaluation protocol, we assess representative video generation systems under both prompt control levels. 
Videos up to $10$s are generated in a single pass, while longer videos of $10$–$45$s are produced through agent-mode workflows when available. 
The results reveal a persistent trade-off between keyframe fidelity and natural video synthesis. 
Models that adhere closely to the input keyframes often connect them with abrupt or implausible transitions, whereas models that use the keyframes as looser semantic guidance tend to drift from their visual content. 
No evaluated system performs best on both axes.
Controllability further deteriorates as the constraints become denser. 
Instruction adherence declines with increasing keyframe count, and explicit camera-motion descriptions rarely yield genuine one-take continuity. 
Moreover, reliable multi-keyframe comprehension remains a major barrier for open-source systems: even the strongest open-source model trails the leading proprietary system by $0.153$ in overall score.

Our contributions are summarized as follows:

\begin{itemize}[leftmargin=*]

\item We introduce \textbf{\name}, the first benchmark dedicated to multi-keyframe-conditioned video generation, comprising $386$ curated test cases with systematically controlled evaluation factors.

\item We decompose keyframe execution into six measurable metrics covering \textit{presence}, \textit{appearance fidelity}, \textit{temporal placement}, \textit{ordering}, \textit{persistence}, and \textit{uniqueness}.

\item We design an evidence-grounded quality evaluation framework that combines MLLM judgments with specialized perception models.

\item We provide stratified analyses of representative video generation systems, revealing systematic performance patterns and failure modes across benchmark settings.

\end{itemize}
\section{Related Work}
\label{sec:related}

\begin{table*}[htbp]
\centering
\caption{\textbf{Comparisons between \name and relative image-conditioned video generation benchmarks.} ``--'' indicates that the aspect is not reported or not used as an explicit benchmark dimension.}
\label{tab:related_benchmark_compare}
\footnotesize
\setlength{\tabcolsep}{3.2pt}
\renewcommand{\arraystretch}{1.10}
\resizebox{\textwidth}{!}{%
\begin{tabular}{@{}lcccccccc@{}}
\toprule
\makecell[l]{\textbf{Benchmark}}
&
\makecell[c]{\textbf{\#Samples}}
&
\makecell[c]{\textbf{\#Metrics}}
&
\makecell[c]{\textbf{Video}\\\textbf{Structure}}
&
\makecell[c]{\textbf{Multi-granular}\\\textbf{Prompt Control}}
&
\makecell[c]{\textbf{Visual Input}\\\textbf{Format}}
&
\makecell[c]{\textbf{\#Keyframe}}
&
\makecell[c]{\textbf{Keyframe Response}\\\textbf{Metrics}}
&
\makecell[c]{\textbf{Audio-Visual}\\\textbf{Metrics}} \\
\midrule

AIGCBench~\cite{fan2024aigcbench}
& 3928
& 11
& --
& \xmark
& Single Image
& --
& \xmark
& \xmark \\

UI2V-Bench~\cite{zhang2025ui2vbenchunderstandingbasedimagetovideogeneration}
& $\sim$500
& 19
& --
& \xmark
& Single
& --
& \xmark
& \xmark \\

MuSS~\cite{zhang2026muss}
& 200
& 14
& Multi-shot
& \xmark
& Single Image
& --
& \xmark
& \xmark \\

LongAV-Compass~\cite{liu2026longavcompass}
& 284
& 29
& --
& \cmark
& Single Image / Video
& --
& \xmark
& \cmark \\

VideoCanvasBench~\cite{cai2026videocanvas}
& 2030
& 9
& --
& \xmark
& \makecell[c]{Multi Images /\\Video Clips / Patches}
& 1 / 2 / 3
& \xmark
& \xmark \\

CineBench~\cite{chen2026cinedancenextgenerationmultishotlongform}
& 1000
& 15
& Multi-shot
& \xmark
& --
& --
& \xmark
& \cmark \\

ViStoryBench~\cite{zhuang2026vistorybench}
& 80
& 12
& Multi-shot
& \xmark
& Multi Images
& --
& \xmark
& \xmark \\

MSVBench~\cite{shi2026msvbenchhumanlevelevaluationmultishot}
& 20
& 20
& Multi-shot
& \xmark
& Multi Images
& \makecell[c]{--}
& \xmark
& \xmark \\

MSAVBench~\cite{wei2026msavbenchcomprehensivereliableevaluation}
& 286
& 20
& Multi-shot
& \xmark
& Multi Images
& --
& \xmark
& \cmark \\
\midrule

\textbf{\name}
& \textbf{386}
& \textbf{15}
& \makecell[c]{\textbf{Single-shot}\\\textbf{/ Multi-shot}}
& \textbf{\cmark}
& \makecell[c]{\textbf{Multi Images}\\\textbf{/ Storyboard Grid}}
& \textbf{3 / 6 / 9 / 12}
& \textbf{\cmark}
& \textbf{\cmark} \\

\bottomrule
\end{tabular}}
\end{table*}

\subsection{Image-Conditioned Video Generation}

Image-conditioned video generation extends text-to-video models with visual references to preserve subject appearance, scene layout, and composition. 
While conventional image-to-video methods mainly animate a single input image, recent approaches increasingly adopt storyboards, predefined keyframes~\cite{liu2025dreamontagea,zhang2026stage}, arbitrary frame guidance~\cite{xiao2025captain,zhang2026smartdirector}, and interleaved multimodal inputs~\cite{pan2026omniweaving}. 
These methods provide stronger control over key event states, shot composition, narrative pacing, and cross-shot consistency~\cite{zhou2026videomemory}. 
However, their evaluations still rely largely on generic video quality, reference fidelity, transition smoothness, or cross-shot consistency. 
They do not provide a unified diagnostic assessment of whether an externally provided sequence of keyframes is faithfully realized in the specified order and at the intended temporal locations.

\subsection{Benchmarks for Video Generation}
General video generation benchmarks, including VBench~\cite{huang2023vbench}, EvalCrafter~\cite{liu2023evalcrafter}, TC-Bench~\cite{feng2024tcbenchbenchmarkingtemporalcompositionality}, and VBench-2.0~\cite{zheng2025vbench2}, primarily evaluate perceptual quality, prompt following, temporal constraints, compositionality, and physical plausibility.
Recent benchmarks further extend evaluation to audio-video generation and long-form synthesis. 
VABench~\cite{hua2026vabenchcomprehensivebenchmarkaudiovideo}, T2AV-Compass~\cite{cao2026t2avcompass}, AVGen-Bench~\cite{zhou2026avgenbenchtaskdrivenbenchmarkmultigranular}, LongAV-Compass~\cite{liu2026longavcompass}, and MSVBench~\cite{shi2026msvbenchhumanlevelevaluationmultishot} assess audio-video coordination, long-form coherence, or script-conditioned generation. 
However, they evaluate final outputs rather than whether an ordered sequence of keyframes is faithfully reproduced at the intended temporal locations with coherent transitions.
For image-conditioned video generation, AIGCBench~\cite{fan2024aigcbench} and UI2V-Bench~\cite{zhang2025ui2vbenchunderstandingbasedimagetovideogeneration} mainly evaluate single-image conditioning, while VideoCanvasBench~\cite{cai2026videocanvas} and ViStoryBench~\cite{zhuang2026vistorybench} extend evaluation to arbitrary spatiotemporal completion and story visualization. 
Nevertheless, they remain focused on visual-condition preservation, visual completion, or coherent image sequences rather than ordered keyframe-conditioned video generation.
In contrast, as summarized in Table~\ref{tab:related_benchmark_compare}, \name fills this gap by providing a diagnostic benchmark for ordered keyframe-conditioned video generation, combining keyframe response evaluation with comprehensive video quality assessment.
\section{\name}
\label{sec:benchmark}
\subsection{Benchmark Design}
\label{sec:benchmark:taxonomy}

Each sample in \textbf{\name} defines a keyframe-conditioned video generation task in which a model receives a sequence of keyframe images and a text prompt and generates a video that realizes the specified visual states in the intended temporal order. The model must reproduce each keyframe at its intended location while synthesizing the motion, transitions, and events between consecutive anchors.

We define the intended keyframe locations according to the temporal structure of the video. In a multi-shot video, each keyframe is assigned to a specific shot with one of three positional roles: \textit{first}, \textit{last}, or \textit{representative}. A \textit{first} keyframe appears at the beginning of its assigned shot, a \textit{last} keyframe at the end, and a \textit{representative} keyframe anywhere within the shot. In a one-take video, each keyframe is assigned a target timestamp along the continuous trajectory. Adjacent keyframes delimit temporal segments, and the corresponding segment descriptions specify how the video should evolve between them, including camera motion, subject evolution, and narrative progression.

To characterize these tasks systematically, each sample is annotated along five dimensions: \textit{input format}, \textit{prompt control level}, \textit{video structure}, \textit{keyframe count}, and \textit{application domain}. These metadata support fine-grained, stratified evaluation across different generation settings.

\noindent\textbf{Prompt control levels.}
Each sample includes two prompt variants. The \textit{minimal prompt} specifies only the keyframe order, a brief story synopsis, the target video structure, and the duration, leaving the model to infer the remaining details from the visual sequence. The \textit{segment-specific prompt} additionally specifies the temporal placement, subject states, camera language, and narrative content of each shot or inter-keyframe segment. Comparing the two variants distinguishes generation under weak guidance from instruction following under explicit constraints. Prompt templates are provided in Appendix~\ref{sec:prompt-generation-video}.

\noindent\textbf{Video structure.}
We consider two structures that capture common video production patterns. A \textit{one-take} sample places all keyframes within a single continuous shot, emphasizing motion generation between successive visual states. A \textit{multi-shot} sample distributes keyframes across multiple shots, emphasizing shot transitions, cross-shot consistency, and event continuity.

\noindent\textbf{Keyframe count.}
We consider $3$, $6$, $9$, and $12$ keyframes. Fewer keyframes require the model to infer longer transitions, whereas more keyframes impose denser visual constraints and leave less freedom for generation.

\noindent\textbf{Input formats.}
Each keyframe sequence is provided in two formats. The \textit{multi-image list} presents the keyframes as an ordered sequence of individual images, whereas the \textit{storyboard grid} arranges the same sequence in a single image for models that accept only one visual input. Both formats preserve identical content and ordering, enabling controlled comparison across conditioning interfaces.

\noindent\textbf{Application domain.}
Samples are grouped into three domains. \textit{Daily capture} covers everyday activities and natural state changes. \textit{Product visualization} focuses on object demonstrations and usage scenarios. \textit{Cinematic narrative} emphasizes story-driven generation involving character interactions and cross-shot continuity.

\subsection{Data Construction}
\label{sec:benchmark:construction}

\begin{figure*}[t]
  \centering
  \includegraphics[width=\textwidth]{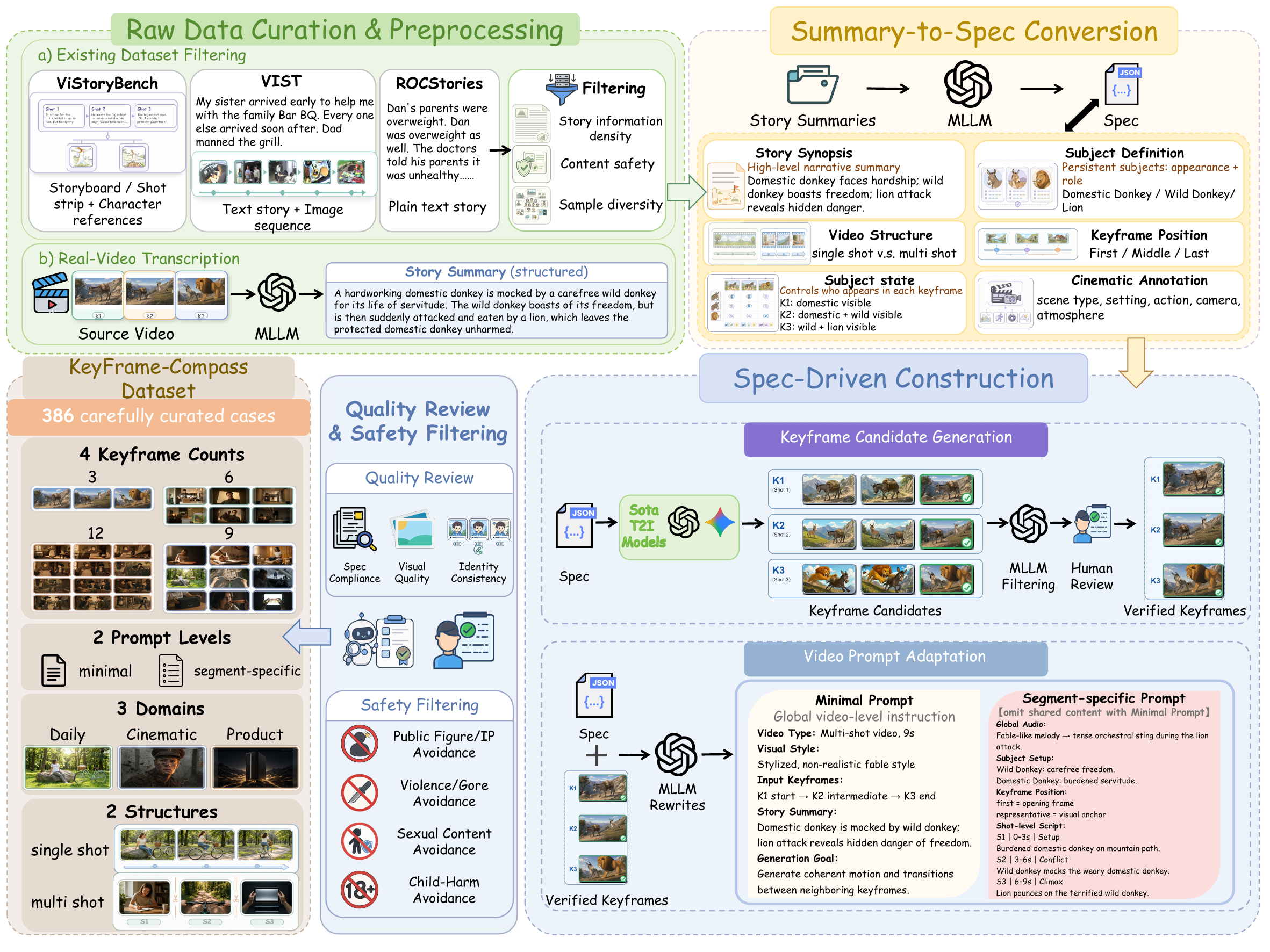}
  \caption{
\textbf{Data construction of \name.} Narrative sources are curated and converted into structured story summaries, which are transformed into scene specifications for keyframe generation and video prompt construction. Candidate samples are then filtered through multimodal verification, human review, and safety screening to ensure quality, consistency, specification compliance, and safety.
  }
  \label{fig:data_construction}
\end{figure*}

\noindent\textbf{Source data.}
As shown in Figure~\ref{fig:data_construction}, \name is constructed from two sources of narrative data. 
The first consists of stories collected from ViStoryBench\cite{zhuang2026vistorybench}, VIST~\cite{tinghao2016visualstorytelling}, and ROCStories~\cite{mostafazadeh2016roc}. 
The second is derived from selected real videos by generating narrative captions that describe their visual content. 
Both sources are converted into structured story summaries, which serve as the basis for subsequent specification construction.

\noindent\textbf{Specification construction.}
For each story summary, we construct a structured scene specification containing the narrative synopsis, video structure, subject descriptions, shot definitions, temporal layout, target keyframe positions, and cinematic annotations. 
This specification serves as the unified representation for keyframe generation, prompt construction, and metadata extraction, ensuring consistent subject identity, scene layout, event ordering, and temporal alignment throughout each sample.

\noindent\textbf{Keyframe image generation.}
Each temporal anchor is translated into image constraints covering subject appearance, action states, spatial layout, and composition. 
Multiple candidate keyframes are generated with GPT-Image-2~\cite{openai2026chatgptimages20} and Nano Banana Pro~\cite{nanobananapro}, then screened through multimodal consistency checks and human review to ensure visual quality, cross-frame subject consistency, narrative relevance, and compliance with the scene specification.

\noindent\textbf{Video prompt adaptation.}
The scene specification is rewritten by Gemini 3.1 Pro~\cite{google2026gemini31pro} into video-oriented prompts while preserving subject identity, visual states, and spatial relations. 
After quality review and safety screening, each benchmark sample consists of an ordered keyframe set in two input formats, two prompt variants, and complete metadata for evaluation.

\subsection{Evaluation Metrics}
\label{sec:benchmark:metric_design}

We evaluate generated videos from two complementary perspectives.
\textit{Keyframe response metrics} measure whether the input keyframes are reproduced with correct fidelity, timing, and temporal structure.
\textit{General quality metrics} assess overall video quality, including visual quality, spatiotemporal coherence, instruction adherence, and audio--visual alignment.

General quality metrics are evaluated with a checklist-based protocol. Given a test case, we use GPT-5.5~\cite{openai2026gpt55} to generate a set of observable items from a metric-specific checklist prompt (App.~\ref{sec:checklists}), defining the evidence to be verified. Gemini 3.1 Pro~\cite{google2026gemini31pro} then scores the generated video against these items following the rubrics in App.~\ref{sec:judge-prompt}. Where applicable, auxiliary model outputs are either supplied to Gemini 3.1 Pro as additional context or fused with its score through metric-specific weights, as detailed below. All scores are normalized to $[0,1]$. 

\begin{figure*}[htbp]
    \centering
    \includegraphics[width=1\linewidth]{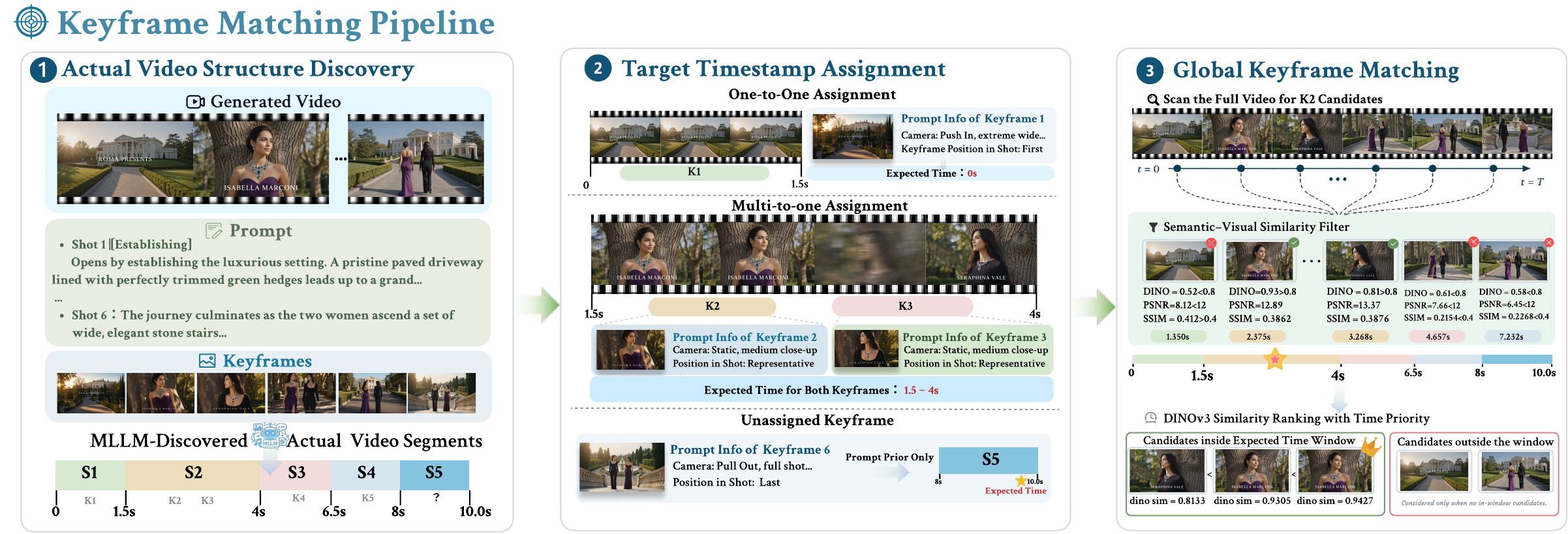}
    \caption{\textbf{Keyframe matching pipeline of \name.}
Since generated videos may not preserve the assumed one-keyframe-per-shot structure, we first use Gemini 3.1 Pro to recover the actual shot structure and assign input keyframes to the corresponding generated segments.
Expected temporal windows are then derived for one-to-one, multi-to-one, and unassigned cases. Within each window, candidate frames are filtered by DINOv3 semantic similarity and PSNR/SSIM pixel fidelity, and the valid candidate with the highest DINOv3 similarity is selected as the canonical match for evaluation.}
    \label{fig:matching}    
\end{figure*}

\paragraph{Keyframe Response Metrics.}
All keyframe response metrics rely on the shared matching pipeline illustrated in Figure~\ref{fig:matching}. Because a generated video may not preserve the assumed one-keyframe-per-shot structure, we first use Gemini 3.1 Pro to infer its actual shot structure, map each input keyframe to the corresponding generated segment, and determine an \textit{expected temporal window} for each assignment case (Appendix~\ref{sec:segmentation}). Within this window, we compare the keyframe against every generated frame using complementary semantic and pixel-level criteria. Following prior work that applies global average pooling to DINO features for image retrieval and alignment~\cite{kabra2026mixeddiet}, we average the spatial patch tokens produced by DINOv3~\cite{simeoni2025dinov3}, $L_2$-normalize the resulting descriptor for each keyframe and video frame, and compute the cosine similarity between them. We retain a candidate frame only if it meets the semantic threshold ($c^{\mathrm{dino}}\geq\tau_d$) and at least one pixel-level threshold (PSNR~$\geq\tau_p$ or SSIM~$\geq\tau_s$). Among all valid candidates, we select the frame with the highest DINOv3 similarity as the canonical match. If no candidate satisfies these criteria, we mark the keyframe as unmatched. Thresholds are calibrated on a held-out, human-annotated validation set, as detailed in Appendix~\ref{app:kfs_details}.

We organize these metrics into two groups: \textbf{Keyframe Fidelity} and \textbf{Keyframe Temporal Organization}.

\textbf{\textit{Keyframe Fidelity.}} This metric group evaluates whether each input keyframe is reproduced and how faithfully it is rendered.

\begin{itemize}[leftmargin=*]

\item \textbf{Hit Rate (HR).}
Hit Rate reports the fraction of input keyframes for which a valid match exists within the expected window (Equation~\ref{eq:hit_rate}), where $N_\mathrm{hit}$ is the number of successfully matched keyframes and $N_\mathrm{key}$ is the total number of input keyframes. This metric captures the basic failure mode of keyframe omission.
\begin{equation}
\label{eq:hit_rate}
\mathrm{HR} = \frac{N_\mathrm{hit}}{N_\mathrm{key}} .
\end{equation}

\item \textbf{Keyframe Similarity (KFS).}
This metric measures the reproduction fidelity of the canonical matched frames identified by the pipeline above. For
each matched pair, we compute a pixel-level similarity from PSNR and SSIM,
normalize it to $[0,1]$, and combine it with the DINOv3 similarity:
\begin{equation}
\label{eq:keyframe_similarity}
\begin{aligned}
q_i
&=
w_{\mathrm{pix}}\,s_i^{\mathrm{pix}}
+
w_{\mathrm{dino}}\,c_i^{\mathrm{dino}},
\qquad i\in\mathcal{M}, \\
\mathrm{KFS}
&=
\frac{1}{N_{\mathrm{kf}}}
\sum_{i\in\mathcal{M}} q_i ,
\end{aligned}
\end{equation}
where $w_{\mathrm{pix}}=0.4$, $w_{\mathrm{dino}}=0.6$, and
$N_{\mathrm{kf}}$ is the total number of input keyframes. 

\end{itemize}

\textbf{\textit{Keyframe Temporal Organization.}}
This metric group evaluates whether reproduced keyframes respect the intended temporal order, appear at appropriate positions, persist naturally, and remain uniquely placed in the generated video. All metrics in this group are computed only over matched keyframes, considering missing responses are already penalized by ~\textit{Hit Rate}.

\begin{itemize}[leftmargin=*]

\item \textbf{Keyframe Position Accuracy (KPA).}
We distinguish two temporal roles for matched keyframes. For \textit{point-position} keyframes, which are specified either by a shot-relative position (\textit{first} or \textit{last}) in multi-shot videos or by an explicit timestamp in one-take videos, the score decays linearly as the matched timestamp $t_i^\mathrm{match}$ deviates from the target $t_i^\mathrm{target}$, with tolerance $\epsilon_i$ set to the maximum distance from the target to either boundary of the expected window $[t_i^\mathrm{start}, t_i^\mathrm{end}]$. 
For keyframes with a \textit{representative} role, the prompt only specifies the shot or segment in which the keyframe should appear, without requiring a specific position within that shot. We apply a binary presence rule: the score is $1$ if the matched timestamp falls inside the assigned segment or expected window, and $0$ otherwise. 
Let $\mathcal{P}$ and $\mathcal{R}$ denote the point-position and representative sets among $N_\mathrm{match}$ matched keyframes.
\begin{equation}
\label{eq:keyframe_position}
\begin{split}
p_i &= \begin{cases}
\max\!\left(0,\;1-\dfrac{|t_i^\mathrm{match}-t_i^\mathrm{target}|}{\epsilon_i}\right), & i\in\mathcal{P}, \\[6pt]
\mathbf{1}\!\left[t_i^\mathrm{match}\in[t_i^\mathrm{start},t_i^\mathrm{end}]\right], & i\in\mathcal{R},
\end{cases}\\[6pt]
\mathrm{KPA} &= \frac{1}{N_\mathrm{match}}\sum_{i=1}^{N_\mathrm{match}} p_i .
\end{split}
\end{equation}

\item \textbf{Keyframe Order Consistency (KOC).}
To assess whether the reproduced keyframes follow the prescribed temporal order, we identify the best-matching frame for each keyframe across the \textit{entire} video, without applying temporal window constraints, and compute Kendall's $\tau$ between the input keyframe order and their matched timestamps. Only keyframes with at least one valid match are included.
Equation~\ref{eq:keyframe_order} normalizes $\tau$ to $[0,1]$:
\begin{equation}
\label{eq:keyframe_order}
\mathrm{KOC} =
\frac{\tau + 1}{2}.
\end{equation}

\item \textbf{Persistence Around Keyframe (PAK).}
This metric detects two temporal failure modes: a keyframe appearing only briefly (\textit{flash}) and remaining as a near-static frame for too long (\textit{freeze}). Within the expected window, we extract the continuous response interval associated with each matched keyframe. For a \textit{representative} keyframe, we select the contiguous response interval containing the canonical match; for a \textit{first} or \textit{last} keyframe, we select an interval anchored near the beginning or end of its expected window, respectively. We assign a flash score $q_i^{\mathrm{flash}}$ based on response duration and a freeze score $q_i^{\mathrm{freeze}}$ based on temporal variation. PAK takes the lower score for each keyframe, penalizing either failure mode:
\begin{equation}
\label{eq:keyframe_persistence}
\mathrm{PAK}
= \frac{1}{N_{\mathrm{match}}}
\sum_{i=1}^{N_{\mathrm{match}}}
\min\!\left(q_i^{\mathrm{flash}},q_i^{\mathrm{freeze}}\right).
\end{equation}
A high PAK therefore requires the keyframe response to persist for sufficient time while preserving temporal dynamics.

\item \textbf{Response Uniqueness (RU).}
We measure whether each keyframe is reproduced in a single coherent temporal region rather than repeatedly appearing at disconnected locations. For each keyframe, we collect all matching frames across the full video and cluster them by temporal proximity. Matches separated by at most
$g=\max(g_{\mathrm{floor}},\,T/N_{\mathrm{kf}}\cdot r)$
belong to the same cluster, where $T$ is the video duration, $N_{\mathrm{kf}}$ is the number of keyframes, and $r$ controls the clustering gap. Let $n_c$ denote the number of matches in cluster $c$. A single response cluster receives a score of $1$; otherwise, the score is the fraction of matches contained in the dominant cluster:
\begin{equation}
\label{eq:response_uniqueness}
\begin{split}
u_i &= \begin{cases}
1 & \text{if one cluster,}\\[2pt]
\dfrac{\max_c\, n_c}{\sum_c n_c} & \text{otherwise,}
\end{cases}\\[6pt]
\mathrm{RU} &= \frac{1}{N_\mathrm{resp}}
\sum_{i=1}^{N_\mathrm{resp}} u_i .
\end{split}
\end{equation}

\end{itemize}

These six metrics jointly diagnose whether keyframes are present, faithful, correctly ordered, accurately timed, naturally sustained, and uniquely placed.

\paragraph{General Quality Metrics.}

General quality metrics are organized into four dimensions: \textbf{Video Quality}, \textbf{Spatiotemporal Coherence}, \textbf{Instruction Adherence}, and \textbf{Audio-Visual Coordination}. Each dimension comprises several metrics defined below. 

\textbf{\textit{Video Quality.}}
We evaluate video quality from both static and temporal perspectives, as high-quality individual frames do not necessarily imply smooth and artifact-free motion.

\begin{itemize}[leftmargin=*]

\item \textbf{Static Visual Quality.}
We assess image fidelity along four axes: clarity and detail preservation, rendering artifacts (e.g., noise, banding, and aliasing), color and lighting consistency, and style consistency across frames. Video-level DOVER~\cite{wu2023dover} and frame-level MUSIQ~\cite{ke2021musiq} scores provide auxiliary evidence.

\item \textbf{Dynamic Visual Quality.}
Temporal quality is assessed separately, covering motion discontinuities, stutter and repeated frames, and temporal flickering. Video-level DOVER scores provide auxiliary evidence.

\end{itemize}

\textbf{\textit{Spatiotemporal Coherence.}}
This metric group evaluates whether the generated video remains coherent over time and physically plausible within the depicted scene.

\begin{itemize}[leftmargin=*]

\item \textbf{Attribute Consistency.}
We evaluate whether foreground entities and backgrounds preserve their visual and semantic attributes throughout the video. For each character, animal, or prop specified in the evaluation checklist, SAM~3.1~\cite{meta2026sam31} tracks the entity across the video and produces temporally aligned masks; backgrounds are evaluated at the full-frame level. The tool-based score uses category-specific features: full-frame DINOv3 embeddings for backgrounds, masked-crop DINOv3 embeddings for animals and props, and specialized appearance embeddings for characters. Specifically, realistic characters are evaluated using ElasticFace~\cite{boutros2022elasticface} for facial identity and InceptionNeXt~\cite{yu2025inceptionnext} for body appearance, while animated characters use an anime-specialized CLIP model. Gemini 3.1 Pro complements these features by assessing semantic attributes such as clothing, color, object state, and other category-specific details.

For each evaluated category $c$, the tool-based and semantic scores are equally weighted:
\[
S_\mathrm{attr}^{(c)}
= 0.5 S_\mathrm{tool}^{(c)} + 0.5 S_\mathrm{Gemini}^{(c)}.
\]
The final Attribute Consistency score is averaged across all evaluated categories.

\item \textbf{Spatial Orientation Consistency.}
Subjects should preserve consistent spatial positions and relations after camera motion is accounted for. Following MSVBench~\cite{shi2026msvbenchhumanlevelevaluationmultishot}, we convert MonST3R~\cite{zhang2025monst3r} frame-wise 6-DoF camera trajectories into semantic camera-motion descriptors, such as ``tilt up''. These descriptors are provided to Gemini 3.1 Pro as auxiliary context, which allows the assessment to distinguish true subject position drift from apparent displacement caused by camera motion.

\item \textbf{Physical Rationality.}
This metric evaluates whether the generated video follows basic physical and commonsense constraints. Gemini 3.1 Pro scores the video along five dimensions: interaction plausibility, including contact, occlusion, grasping, and collision; anatomical consistency for biological subjects; structural stability; gravity and material behavior; and scene-level common sense. The final Physical Rationality score is computed as the average of the five dimension scores.

\end{itemize}

\textbf{\textit{Instruction Adherence.}}
To evaluate models' compliance with input instructions, we devise two separate metrics targeting the video modality and audio modality respectively.

\begin{itemize}[leftmargin=*]

\item \textbf{Video Modality Adherence.}
This metric is evaluated under both prompt modes and measures how closely the generated video follows the structured directives. We group the evaluation criteria into four categories: (1) \textit{camera execution}, which assesses whether the realized shot scale and camera movement match the detailed camera instructions. MonST3R-derived semantic motion descriptors are provided to Gemini 3.1 Pro as auxiliary context to determine whether the realized camera behavior matches the requested specification.; (2) \textit{shot structure}, which checks whether the output follows the requested multi-shot or one-take format; (3) \textit{narrative pacing}, which evaluates whether the timing and emphasis of the video reflect the intended focus of the story; and (4) \textit{subject--scene alignment}, which verifies that each required subject appears in the correct scene. The exact evaluation scope and decision boundaries are defined separately for each mode; detailed definitions are provided in the Appendix~\ref{sec:checklists}.

\item \textbf{Audio Modality Adherence.}
This metric applies to segment-specific prompts with explicit audio requirements. It evaluates whether the generated audio matches the requested sound content, style, and emotional tone and whether audio events synchronize with the corresponding visual events.

Gemini 3.1 Pro provides the checklist-based
score~\(S_{\scriptscriptstyle\mathrm{Gemini}}^{\scriptscriptstyle\mathrm{audio}}\). As an auxiliary signal, we compute the CLAP audio--text similarity between the generated audio $A$ and the audio requirement text $T_\mathrm{audio}$:
\begin{equation}
\label{eq:audio_clap}
C_{\scriptscriptstyle\mathrm{CLAP}}^{\scriptscriptstyle\mathrm{audio}}
= \cos\!\left(E_a(A),\; E_t(T_\mathrm{audio})\right),
\end{equation}
where $E_a$ and $E_t$ are the CLAP audio and text encoders, respectively. 

Since CLAP measures global audio--text alignment but cannot verify event timing or lip synchronization, the final score combines the Gemini score with the normalized CLAP signal:
\begin{equation}
S_\mathrm{audio}
= \lambda_\mathrm{audio} S_{\scriptscriptstyle\mathrm{Gemini}}^{\scriptscriptstyle\mathrm{audio}}
+ (1-\lambda_\mathrm{audio}) \tilde{C}_{\scriptscriptstyle\mathrm{Gemini}}^{\scriptscriptstyle\mathrm{audio}},
\end{equation}
where $\lambda_\mathrm{audio}$ is the predefined fusion weight.

\end{itemize}

\textbf{\textit{Audio-Visual Coordination.}}
For samples with audio requirements, we additionally evaluate audio-visual agreement independent of prompt adherence. We report all cosine-based audio--visual scores using the normalized cosine similarity
$s(x,y)=\frac{1}{2}(\cos(x,y)+1)\in[0,1]$, where higher is better.

\begin{itemize}[leftmargin=*]

\item \textbf{ImageBind Similarity.}
We measure global audio--visual semantic consistency in the ImageBind~\cite{girdhar2023imagebind} embedding space:
\begin{equation}
\label{eq:imagebind_av}
\mathrm{IB}_{\mathrm{av}}
=
s\!\left(
\operatorname{Pool}_{t} E_v(v_t),
E_a(A)
\right),
\end{equation}
where $A$ is the generated audio, $v_t$ is the $t$-th sampled video frame, and $E_a,E_v$ are the ImageBind audio and visual encoders.

\item \textbf{JavisScore.}
Following JavisDiT~\cite{liu2025javisdit}, we measure window-level audio--visual synchrony by averaging the least aligned frames in each temporal window:
\begin{equation}
\label{eq:javis_score}
\mathrm{JavisScore}
=
\frac{1}{M}
\sum_{i=1}^{M}
\frac{1}{|\mathcal{B}_i|}
\sum_{j\in\mathcal{B}_i}
s\!\left(E_v(v_{i,j}),E_a(a_i)\right),
\end{equation}
where $a_i$ is the audio segment in the $i$-th window, $v_{i,j}$ is the corresponding $j$-th video frame, and $\mathcal{B}_i$ contains the lowest-scoring 40\% frames in that window.

\end{itemize}

\subsection{Statistics}
\label{sec:benchmark:statistics}

\begin{figure*}[htbp]
  \centering
  \includegraphics[width=\textwidth]{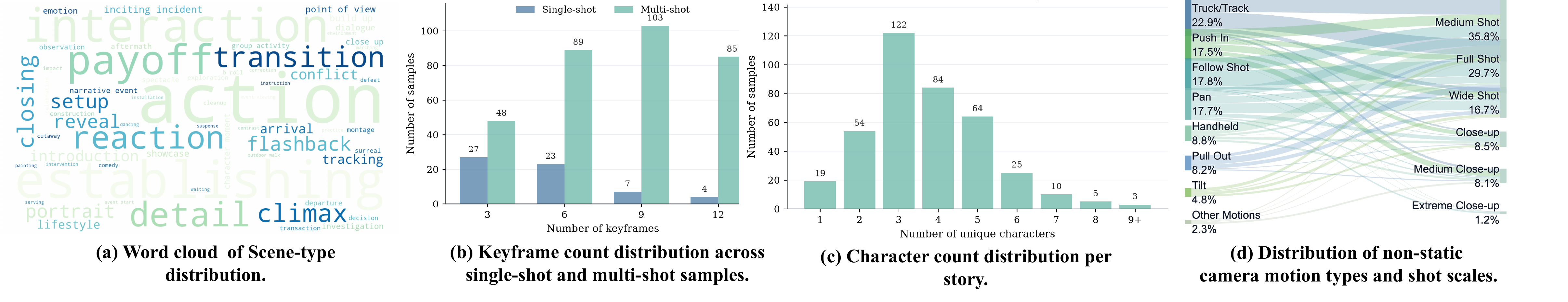}
  \caption{
  \textbf{Dataset statistics of \name.} From left to right: the scene-
  type word cloud, the distribution of characters
  per story, the distribution of keyframe counts across one-take and multi-shot samples, and the co-occurrence patterns between camera motions and shot scales. These statistics summarize the diversity of story structure, scene
  semantics, and camera-control requirements in the benchmark. 
  }
  \label{fig:panel}
\end{figure*}

\begin{table}[htbp]
\centering
\caption{\textbf{Distribution of samples in \name across video structure, target duration, and application domain.}}
\label{tab:data_distribution}
\footnotesize
\setlength{\tabcolsep}{0pt}
\renewcommand{\arraystretch}{1.06}
\begin{tabular*}{\linewidth}{@{\extracolsep{\fill}}cccccc@{}}
\toprule
\multicolumn{2}{c}{\textbf{Video Setting}} &
\multicolumn{3}{c}{\textbf{Application Domain}} &
\multicolumn{1}{c}{\textbf{Total}} \\
\cmidrule(lr){1-2}\cmidrule(lr){3-5}\cmidrule(l){6-6}
\textbf{Structure} &
\textbf{Duration} &
\textbf{Daily} &
\textbf{Product} &
\textbf{Cinematic} &
\textbf{Cases} \\
\midrule
One-Take & 5--10\,s & 14 & 4 & 43 & 61 \\
Multi-Shot & 5--10\,s & 110 & 32 & 24 & 166 \\
Multi-Shot & 10--45\,s & 66 & 17 & 76 & 159 \\
\midrule
\multicolumn{2}{c}{\textbf{Total}} & \textbf{190} & \textbf{53} & \textbf{143} & \textbf{386} \\
\bottomrule
\end{tabular*}
\end{table}

The final benchmark comprises $386$ test cases spanning three video structure-duration settings and three application domains, as summarized in Table~\ref{tab:data_distribution}. 
Specifically, it covers one-take videos of $5$–$10$s, short multi-shot videos of $5$–$10$s, and long multi-shot videos of $10$–$45$s. 
Figure~\ref{fig:panel} further presents the detailed composition of \name. 
On average, each story contains $3.73$ unique characters, with most stories involving three to $5$ characters. 
The benchmark includes $75$, $112$, $110$, and $89$ samples with $3$, $6$, $9$, and $12$ keyframes, respectively. 
Scene annotations are dominated by \textit{action}, \textit{establishing}, \textit{interaction}, and \textit{payoff}, which together account for $67.2$\% of all scene labels. 
We further analyze shot-level camera-language annotations after excluding static-camera cases to better characterize the diversity of intentional camera control. 
Among the $730$ annotated motion–scale pairs, truck/track, follow shot, pan, and push-in are the most common camera motions, while medium, full, and wide are the predominant shot scales.

\section{Experiment}
\label{sec:experiments}

\subsection{Experimental Settings}
\label{sec:experiments:settings}

\begin{table*}[htbp]
\centering
\caption{\textbf{Short-video leaderboard of joint audio-video generation models on \name.}
Scores are based on 115 base samples evaluated by all six models in both prompt modes. Dimension scores are averaged across modes; Keyframe Fidelity and Temporal Organization are multiplied before group averaging within each mode, and Overall averages the two mode-level scores. Best and second-best results are \textbf{bolded} and \underline{underlined}, respectively.}
\label{tab:leaderboard}
\scriptsize
\setlength{\tabcolsep}{4pt}
\renewcommand{\arraystretch}{1.12}
\resizebox{\textwidth}{!}{%
\begin{tabular}{@{}r l ccccccc@{}}
\toprule
\textbf{Rank} & \textbf{Model}
& \makecell{\textbf{Keyframe}\\\textbf{Fidelity}}
& \makecell{\textbf{Temporal}\\\textbf{Organization}}
& \makecell{\textbf{Video}\\\textbf{Quality}}
& \makecell{\textbf{Spatiotemporal}\\\textbf{Coherence}}
& \makecell{\textbf{Instruction}\\\textbf{Adherence}}
& \makecell{\textbf{AV}\\\textbf{Coordination}}
& \textbf{Overall} \\
\midrule
1 & Seedance 2.0 & \underline{0.807} & \underline{0.859} & \underline{0.850} & \textbf{0.935} & \textbf{0.931} & \underline{0.626} & \textbf{0.807} \\
2 & Gemini-Omni-Flash & 0.483 & 0.807 & \textbf{0.861} & \underline{0.905} & \underline{0.923} & \textbf{0.640} & \underline{0.744} \\
3 & Kling-3.0-Omni & 0.665 & 0.805 & 0.813 & 0.876 & 0.871 & 0.598 & 0.738 \\
4 & LTX-2.3 & \textbf{0.855} & \textbf{0.899} & 0.557 & 0.680 & 0.721 & 0.570 & 0.659 \\
5 & Wan2.7-I2V & 0.490 & 0.667 & 0.734 & 0.781 & 0.706 & 0.593 & 0.628 \\
6 & daVinci-MagiHuman-1080p-I2V & 0.295 & 0.627 & 0.212 & 0.292 & 0.152 & 0.577 & 0.284 \\
\bottomrule
\end{tabular}
}
\end{table*}

\paragraph{Evaluated Models.}
We evaluate $9$ representative state-of-the-art video generation models on \name, assessing their ability to reproduce the input keyframes at the intended temporal positions while maintaining overall video quality.
Evaluation is performed separately by target duration.
For videos of $10$s or shorter, we evaluate $4$ proprietary models, Gemini-Omni-Flash~\cite{googledeepmind2026geminiomniflash}, Kling-3.0-Omni~\cite{team2025kling}, Seedance~2.0~\cite{seedance2026seedance}, and Wan2.7-I2V~\cite{alibabacloud2026wan27i2v}, together with $5$ open-source models, daVinci-MagiHuman-1080p-I2V~\cite{siigair2026davinci}, HunyuanVideo1.5-I2V~\cite{wu2025hunyuanvideo15}, LTX-2.3~\cite{hacohen2024ltx}, SkyReels-V2-I2V~\cite{chen2025skyreelsv2}, and Wan2.2-I2V-A14B~\cite{wan2025wan}. 
All models are evaluated in a single-pass setting.
For longer videos, we evaluate only Kling-3.0-Omni and Seedance~2.0 using their agent-mode workflows, which generate the final video through multiple sequential passes, since the remaining models do not support long-form generation in our evaluation setup.

\paragraph{Implementation Details.}
Within each duration split, all eligible models are evaluated on the same test cases under both prompt variants.
The input format is adapted to each model's visual conditioning interface: multi-image models receive a multi-image list, while single-image models receive a storyboard grid.
We follow each model's official or recommended inference settings whenever available.
All models generate videos at $720$P except daVinci-MagiHuman-1080p-I2V, which outputs $1080$P. The aspect ratio is determined from the input keyframes and adjusted to the closest ratio supported by each model.

\paragraph{Evaluation Protocol.}
All generated videos are evaluated using the unified pipeline described in Section~\ref{sec:benchmark:metric_design}, with results reported separately for the two prompt variants.
Since minimal prompts do not specify temporal positions or audio requirements, Keyframe Position Accuracy (KPA) and Audio Modality Adherence (AMA) are reported only for the segment-specific prompt.

\subsection{Main Results}
\label{sec:experiments:results}

\newcommand{\tbdres}{\todo{--}}
\newcommand{\nares}{--}

On the common set of $115$ samples used for the joint audio--video leaderboard (Table~\ref{tab:leaderboard}), Seedance~2.0 ranks first with an Overall score of $0.807$, exceeding Gemini-Omni-Flash by $0.063$. Its advantage comes from balanced performance: it ranks first in Spatiotemporal Coherence and Instruction Adherence and second in the other four dimensions. By contrast, LTX-2.3 ranks first in Keyframe Fidelity ($0.855$) and Temporal Organization ($0.899$) but only fourth overall ($0.659$), owing to lower General Video Quality. Gemini-Omni-Flash exhibits the reverse profile, leading in Video Quality ($0.861$) and AV Coordination ($0.640$) despite a Keyframe Fidelity score of $0.483$. Thus, ranking models by keyframe fidelity alone would favor LTX-2.3, ranking them by video quality would favor Gemini-Omni-Flash, whereas the joint evaluation favors Seedance~2.0. These rank reversals demonstrate that input preservation, temporal organization, and general generation quality are non-interchangeable capabilities.

For aggregation, KPA and RU form one component within Keyframe Temporal Organization: $\mathrm{KPA}\times\mathrm{RU}$ under segment-specific prompts and RU alone under minimal prompts, where KPA is undefined. This component is averaged with the remaining metrics in the group. Within each prompt mode, Keyframe Fidelity and Keyframe Temporal Organization are multiplied into a joint keyframe-control contribution before being averaged with the other applicable metric groups. The final Overall score averages the two mode-level scores.

Table~\ref{tab:detailed_results} presents the full segment-specific and minimal prompt results, grouped by video duration.
To ensure fair comparison among proprietary models, short-video scores are aggregated over the sample intersection common to all proprietary systems; open-source models are scored on their full eligible sets. 
We separate the short and long splits because they test fundamentally different generation regimes: single-pass output versus agent-mode multi-pass workflows.

\begin{table*}[t]
\centering
\caption{\textbf{Per-metric results on \name, grouped by video duration.} Each subtable reports segment-specific and minimal prompts as separate row blocks; KPA and AMA are not applicable to minimal prompts. Aud. indicates whether generated audio is available, and audio-only metrics are not applicable to silent models. Colored cells \ranklegend~indicate the first, second, third, and fourth performance within each short-video prompt mode, respectively; \textbf{bold} indicates the better long-video result.}
\label{tab:detailed_results}
\label{tab:specific_results}
\label{tab:minimal_results}

\begin{subtable}{\textwidth}
\centering
\caption{Short-video results. This split contains samples of 10 seconds or shorter, for which every evaluated model produces a single-pass output.}
\label{tab:short_results}
\label{tab:specific_short_results}
\label{tab:minimal_short_results}
\scriptsize
\setlength{\tabcolsep}{2.6pt}
\renewcommand{\arraystretch}{1.05}
\resizebox{\textwidth}{!}{%
\begin{tabular}{@{}l c *{6}{c} *{9}{c}@{}}
\toprule
\textbf{Model} & \textbf{Aud.}
& \multicolumn{6}{c}{Keyframe-specific metrics}
& \multicolumn{9}{c}{General quality metrics} \\
\cmidrule(lr){3-8} \cmidrule(l){9-17}
& & \multicolumn{2}{c}{Fidelity}
& \multicolumn{4}{c}{Temporal Organization }
& \multicolumn{2}{c}{Video Quality}
& \multicolumn{3}{c}{\makecell{Spatiotemporal\\Coherence}}
& \multicolumn{2}{c}{Instruction}
& \multicolumn{2}{c}{AV coord.} \\
\cmidrule(lr){3-4} \cmidrule(lr){5-8} \cmidrule(lr){9-10} \cmidrule(lr){11-13} \cmidrule(lr){14-15} \cmidrule(l){16-17}
& & HR & KFS & KOC & PAK & RU & KPA & SVQ & DVQ & AC & SOC & PR & VMA & AMA & IB & Jav. \\
\midrule
\multicolumn{17}{@{}l}{\textit{Segment-specific prompt}} \\
\midrule
\ClosedGroup{17}
\midrule
Seedance 2.0 & Yes & \ranktwo{0.961} & \ranktwo{0.652} & \ranktwo{0.984} & \rankfour{0.730} & \rankone{0.953} & \ranktwo{0.717} & \ranktwo{0.862} & \ranktwo{0.844} & \rankone{0.967} & \rankone{0.940} & \rankone{0.919} & \ranktwo{0.907} & \rankone{0.884} & \ranktwo{0.656} & \ranktwo{0.595} \\
Gemini-Omni-Flash & Yes & \rankfour{0.623} & \rankfour{0.371} & \rankfour{0.901} & \rankthree{0.798} & 0.922 & 0.587 & \rankone{0.864} & \rankone{0.861} & \rankthree{0.940} & \ranktwo{0.914} & \ranktwo{0.869} & \rankone{0.907} & \ranktwo{0.855} & \rankone{0.671} & \rankone{0.613} \\
Kling-3.0-Omni & Yes & \rankthree{0.778} & \rankthree{0.587} & \rankthree{0.959} & 0.642 & \ranktwo{0.953} & \rankthree{0.659} & \rankthree{0.827} & \rankthree{0.800} & \ranktwo{0.943} & \rankthree{0.883} & \rankthree{0.839} & \rankthree{0.782} & \rankthree{0.855} & \rankfour{0.621} & \rankthree{0.569} \\
Wan2.7-I2V & Yes & 0.614 & 0.368 & 0.748 & 0.670 & \rankthree{0.952} & 0.376 & \rankfour{0.778} & \rankfour{0.688} & \rankfour{0.926} & \rankfour{0.723} & \rankfour{0.629} & 0.523 & \rankfour{0.746} & \rankthree{0.624} & \rankfour{0.567} \\
\midrule
\OpenGroup{17}
\midrule
LTX-2.3 & Yes & \rankone{0.967} & \rankone{0.735} & \rankone{0.985} & \rankone{0.891} & 0.904 & \rankone{0.873} & 0.665 & 0.518 & 0.879 & 0.698 & 0.515 & \rankfour{0.637} & 0.633 & 0.595 & 0.551 \\
Wan2.2-I2V-A14B & No & 0.191 & 0.277 & 0.494 & \ranktwo{0.890} & 0.884 & 0.503 & 0.693 & 0.597 & 0.790 & 0.351 & 0.274 & 0.092 & \nares & \nares & \nares \\
SkyReels-V2-I2V & No & 0.272 & 0.368 & 0.475 & 0.708 & 0.944 & \rankfour{0.596} & 0.406 & 0.374 & 0.635 & 0.156 & 0.160 & 0.102 & \nares & \nares & \nares \\
daVinci-MagiHuman-1080p-I2V & Yes & 0.270 & 0.333 & 0.477 & 0.726 & 0.888 & 0.561 & 0.201 & 0.218 & 0.576 & 0.088 & 0.029 & 0.086 & 0.511 & 0.599 & 0.566 \\
HunyuanVideo 1.5-I2V & No & 0.232 & 0.285 & 0.559 & 0.621 & \rankfour{0.950} & 0.398 & 0.466 & 0.369 & 0.699 & 0.121 & 0.102 & 0.104 & \nares & \nares & \nares \\
\midrule
\multicolumn{17}{@{}l}{\textit{Minimal prompt}} \\
\midrule
\ClosedGroup{17}
\midrule
Seedance 2.0 & Yes & \ranktwo{0.967} & \ranktwo{0.652} & \ranktwo{0.970} & \ranktwo{0.833} & \ranktwo{0.958} & \nares & \ranktwo{0.850} & \ranktwo{0.835} & \rankone{0.963} & \rankone{0.938} & \rankone{0.882} & \rankone{0.970} & \nares & \ranktwo{0.654} & \ranktwo{0.598} \\
Gemini-Omni-Flash & Yes & 0.601 & 0.353 & \rankfour{0.913} & \rankfour{0.778} & 0.922 & \nares & \rankone{0.865} & \rankone{0.857} & \rankthree{0.942} & \ranktwo{0.896} & \ranktwo{0.856} & \ranktwo{0.967} & \nares & \rankone{0.666} & \rankone{0.611} \\
Kling-3.0-Omni & Yes & \rankthree{0.754} & \rankthree{0.572} & \rankthree{0.959} & 0.702 & \rankthree{0.957} & \nares & \rankthree{0.811} & \rankthree{0.802} & \ranktwo{0.948} & \rankthree{0.877} & \rankthree{0.797} & \rankthree{0.925} & \nares & \rankthree{0.625} & \rankthree{0.573} \\
Wan2.7-I2V & Yes & \rankfour{0.606} & 0.362 & 0.751 & 0.494 & 0.937 & \nares & 0.750 & \rankfour{0.716} & \rankfour{0.917} & \rankfour{0.814} & \rankfour{0.680} & 0.783 & \nares & \rankfour{0.618} & \rankfour{0.565} \\
\midrule
\OpenGroup{17}
\midrule
LTX-2.3 & Yes & \rankone{0.979} & \rankone{0.750} & \rankone{0.996} & \rankone{0.882} & 0.883 & \nares & 0.569 & 0.453 & 0.834 & 0.620 & 0.417 & \rankfour{0.801} & \nares & 0.581 & 0.542 \\
Wan2.2-I2V-A14B & No & 0.190 & 0.282 & 0.519 & \rankthree{0.808} & 0.885 & \nares & \rankfour{0.765} & 0.602 & 0.864 & 0.332 & 0.350 & 0.003 & \nares & \nares & \nares \\
SkyReels-V2-I2V & No & 0.264 & \rankfour{0.447} & 0.510 & 0.676 & 0.924 & \nares & 0.566 & 0.445 & 0.826 & 0.339 & 0.381 & 0.010 & \nares & \nares & \nares \\
daVinci-MagiHuman-1080p-I2V & Yes & 0.273 & 0.363 & 0.536 & 0.564 & \rankone{0.972} & \nares & 0.215 & 0.227 & 0.754 & 0.176 & 0.106 & 0.008 & \nares & 0.584 & 0.560 \\
HunyuanVideo 1.5-I2V & No & 0.193 & 0.285 & 0.513 & 0.635 & \rankfour{0.948} & \nares & 0.468 & 0.353 & 0.768 & 0.069 & 0.084 & 0.001 & \nares & \nares & \nares \\
\bottomrule
\end{tabular}
}
\end{subtable}

\vspace{0.6em}

\begin{subtable}{\textwidth}
\centering
\caption{Long-video results. This split contains samples longer than 10 seconds and is evaluated only with agent-mode generation workflows for Kling-3.0-Omni and Seedance 2.0.}
\label{tab:long_results}
\label{tab:specific_long_results}
\label{tab:minimal_long_results}
\scriptsize
\setlength{\tabcolsep}{2.6pt}
\renewcommand{\arraystretch}{1.05}
\resizebox{\textwidth}{!}{%
\begin{tabular}{@{}l c *{6}{c} *{9}{c}@{}}
\toprule
\textbf{Model} & \textbf{Aud.}
& \multicolumn{6}{c}{Keyframe-specific metrics}
& \multicolumn{9}{c}{General quality metrics} \\
\cmidrule(lr){3-8} \cmidrule(l){9-17}
& & \multicolumn{2}{c}{Fidelity}
& \multicolumn{4}{c}{Temporal Organization }
& \multicolumn{2}{c}{Video Quality}
& \multicolumn{3}{c}{\makecell{Spatiotemporal\\Coherence}}
& \multicolumn{2}{c}{Instruction}
& \multicolumn{2}{c}{AV coord.} \\
\cmidrule(lr){3-4} \cmidrule(lr){5-8} \cmidrule(lr){9-10} \cmidrule(lr){11-13} \cmidrule(lr){14-15} \cmidrule(l){16-17}
& & HR & KFS & KOC & PAK & RU & KPA & SVQ & DVQ & AC & SOC & PR & VMA & AMA & IB & Jav. \\
\midrule
\multicolumn{17}{@{}l}{\textit{Segment-specific prompt}} \\
\midrule
Seedance 2.0 (agent) & Yes
& 0.741 & 0.472 & \textbf{0.942} & \textbf{0.781} & \textbf{0.926} & \textbf{0.718}
& 0.785 & \textbf{0.808} & \textbf{0.941} & 0.778 & \textbf{0.732} & \textbf{0.735} & \textbf{0.811} & \textbf{0.653} & \textbf{0.597} \\
Kling-3.0-Omni (agent) & Yes
& \textbf{0.876} & \textbf{0.707} & 0.929 & 0.771 & 0.836 & 0.602
& \textbf{0.794} & 0.789 & 0.935 & \textbf{0.826} & 0.730 & 0.735 & 0.491 & 0.634 & 0.582 \\
\midrule
\multicolumn{17}{@{}l}{\textit{Minimal prompt}} \\
\midrule
Seedance 2.0 (agent) & Yes
& \textbf{0.687} & \textbf{0.440} & 0.859 & \textbf{0.801} & 0.895 & \nares
& \textbf{0.745} & 0.714 & \textbf{0.929} & \textbf{0.791} & \textbf{0.630} & 0.886 & \nares & \textbf{0.655} & \textbf{0.594} \\
Kling-3.0-Omni (agent) & Yes
& 0.551 & 0.344 & \textbf{0.880} & 0.710 & \textbf{0.930} & \nares
& 0.725 & \textbf{0.755} & 0.915 & 0.730 & 0.617 & \textbf{0.899} & \nares & 0.638 & 0.576 \\
\bottomrule
\end{tabular}
}
\end{subtable}
\end{table*}

\begin{figure*}[htbp]
    \centering
    \includegraphics[width=\linewidth]{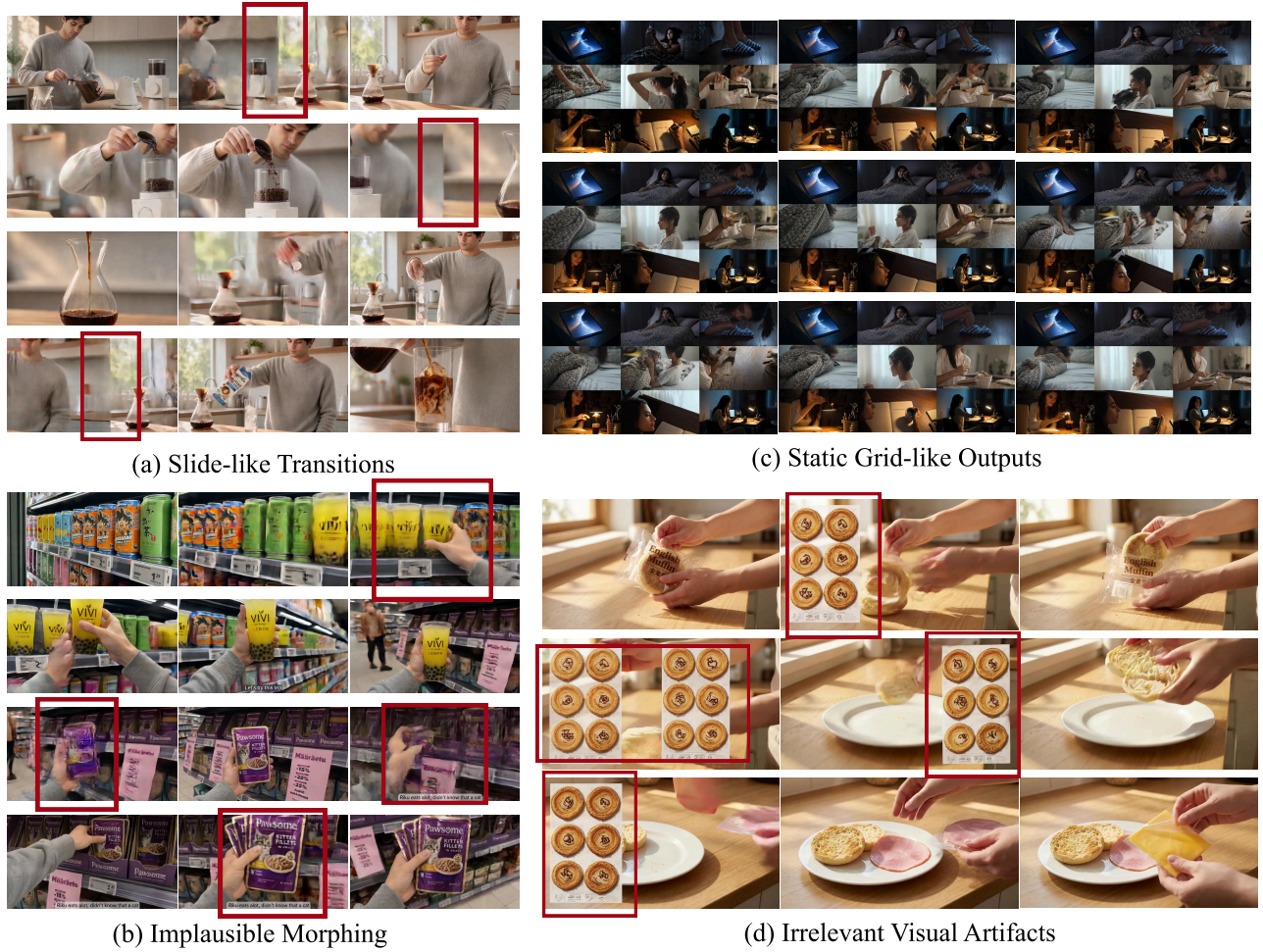}
    \caption{\textbf{Representative failure modes of open-source models.} (a)~\textit{Slide-like transitions} and (b)~\textit{implausible morphing}: LTX-2.3 reproduces the input keyframes faithfully but connects them with abrupt cuts or physically implausible dissolves instead of natural motion. (c)~\textit{Static grid-like outputs}: the storyboard grid is reproduced as a whole rather than decomposed into temporally ordered shots. (d)~\textit{Irrelevant visual artifacts}: content unrelated to the intended shot intrudes into generated frames.}
    \label{fig:open_failure}
\end{figure*}

\subsection{Analysis and Insights}
\label{sec:experiments:analysis}

\paragraph{Strong Keyframe Execution Highlights Transition Synthesis as the Remaining Bottleneck.}
LTX-2.3 achieves the strongest overall performance among the evaluated open-source models and, notably, the highest keyframe response scores among all evaluated systems. 
It reproduces the prescribed keyframes with high fidelity and in the intended temporal order while maintaining competitive overall video quality. 
However, the gap between its strong keyframe adherence and its less consistent temporal dynamics indicates that the model's principal remaining limitation lies in synthesizing plausible transitions between these visual anchors. 
Although most transitions remain coherent, cases involving large changes in scene content, camera viewpoint, or shot scale can produce abrupt, slideshow-like changes, as shown in Figure~\ref{fig:open_failure} (a). 
Similarly, rapid subject motion may lead to melting, dissolving, or fluid-like artifacts in intermediate frames, as illustrated in Figure~\ref{fig:open_failure} (b).
These qualitative failure modes are consistent with its comparatively weaker minimal-mode DVQ ($0.453$) and PR ($0.417$) results, particularly relative to its keyframe response performance. 
Taken together, the qualitative and quantitative evidence suggests that further improvements to LTX-2.3 should focus less on reproducing the specified visual states and more on constructing temporally coherent and physically plausible trajectories between them. 
This distinction also illustrates the diagnostic value of our evaluation framework, which separates successful keyframe realization from the quality of the intervening dynamics.


\paragraph{Semantic compliance does not imply visual grounding to the input.}
A model may satisfy nearly all shot-level instructions while retaining only limited visual correspondence to the provided keyframes. Gemini-Omni-Flash provides the clearest example: with segment-specific prompting, it ranks among the top models on VMA, yet its KFS remains relatively low. As detailed in App.~\ref{closed-cases}, the model often \textit{re-stages} each shot instead of directly continuing the visual content of the input frames. Its outputs retain the high-level cinematic and semantic attributes specified by the prompt and suggested by the keyframe, including shot scale, camera framing, lighting, and subject action. However, scene layout, character appearance, and individual objects may be reconstructed from scratch and thus no longer closely resemble their input counterparts. This behavior reveals a characteristic preference: Gemini-Omni-Flash appears to use keyframes mainly as semantic references for shot content rather than strict visual anchors to reproduce in the final video. Seedance~2.0 shows the opposite preference, using keyframes as stronger appearance anchors. It closely follows both fine-grained instructions and input imagery, but such strict anchoring can limit context-appropriate refinement or completion of the supplied content, occasionally producing less natural transitions between adjacent segments. Kling-3.0-Omni falls between these extremes, preserving moderate keyframe similarity while maintaining comparatively coherent narrative flow. The preferable behavior ultimately depends on how strongly an application requires generated content to remain visually bound to its inputs, a constraint the text prompt alone may not fully specify.

\paragraph{Instruction adherence declines as the density of visual constraints increases.}
The results show a consistent negative association between constraint density and instruction adherence. For all four proprietary models, VMA is lower under segment-specific prompts than under minimal prompts (Tables~\ref{tab:specific_results}), with absolute decreases ranging from $0.060$ to $0.260$. The same pattern is observed within the segment-specific setting as the number of keyframes increases. As shown in Table~\ref{tab:specific_instruction_by_keyframe_count}, the instruction-adherence score decreases from $0.849$ for 3 keyframes to $0.818$ for 6 keyframes and $0.756$ for the merged 9\&12 keyframe group. This decline is driven primarily by VMA, whereas AMA remains nearly unchanged across the three groups. Taken together, these comparisons indicate that the observed degradation is concentrated in visual and temporal control rather than audio adherence. Within a short video, increasing the number of keyframes shortens the intervals between anchors and imposes additional segment-level requirements on camera motion, framing, and temporal placement, while AMA remains stable.

\begin{table}[htbp]
\centering
\caption{Instruction-adherence scores by keyframe count under segment-specific prompts. Instruction averages VMA and AMA; the 9\&12 row merges the two densest keyframe groups.}
\label{tab:specific_instruction_by_keyframe_count}
\scriptsize
\setlength{\tabcolsep}{6pt}
\renewcommand{\arraystretch}{1.05}
\begin{tabular}{@{}lccc@{}}
\toprule
\textbf{Keyframes} & \textbf{Instruction} & \textbf{VMA} & \textbf{AMA} \\
\midrule
3 & 0.849 & 0.861 & 0.837 \\
6 & 0.818 & 0.799 & 0.837 \\
\textbf{9\&12} & \textbf{0.756} & \textbf{0.682} & \textbf{0.830} \\
\bottomrule
\end{tabular}
\end{table}

\paragraph{Open-source models have not acquired storyboard-grid comprehension.}
The gap between proprietary and open-source models on \name reflects a more fundamental difference in their input comprehension. Apart from LTX-2.3, which accepts multi-image conditioning and whose failure modes are analyzed above, the remaining open-source models (daVinci, HunyuanVideo, SkyReels, and Wan2.2-I2V) receive the keyframes as a storyboard grid and fail before generation even begins. Their Hit Rates stay below $0.28$, and their outputs often reproduce the grid as a whole in a near-static video (Figure~\ref{fig:open_failure}c), which suggests that these models have not learned to decompose a grid image into temporally ordered visual conditions. The models also mishandle the conditioning inputs in subtler ways. Generated frames are contaminated by visual artifacts unrelated to the intended shot, such as pasted fragments(Figure~\ref{fig:open_failure}d), and on the audio side, daVinci sometimes reads the prompt text aloud as dialogue or applies the audio requirements of one shot to another. None of these comprehension failures appears in proprietary systems, which points to a gap in training data rather than in model capacity.

\begin{table}[htbp]
\centering
\caption{\textbf{Descriptive model-level rank alignment between human win rates and automatic general-quality scores ($n=5$ models).} Exact $p$-values are computed by two-sided tests over all $5!$ model-label permutations, using average ranks for any tied model-level scores.}
\label{tab:human_alignment}
\scriptsize
\begin{tabular}{@{}lcc@{}}
\toprule
Dimension & $\rho_s$ & Exact $p$ \\
\midrule
Video Quality & 0.90 & 0.083 \\
Spatiotemporal Coherence & 0.80 & 0.133 \\
Instruction Adherence & 1.00 & 0.017 \\
AV Coordination & 0.70 & 0.233 \\
\midrule
Overall & 0.90 & 0.083 \\
\bottomrule
\end{tabular}
\end{table}

\subsection{Human Alignment}
\label{sec:experiments:human-alignment}
We evaluate human alignment on $60$ segment-level cases, stratified by video structure, keyframe count, and content domain, using five models spanning the observed performance range: Seedance~2.0, Gemini-Omni-Flash, Kling-3.0-Omni, Wan2.7-I2V, and LTX-2.3. For each evaluation dimension, we evaluate all $\binom{5}{2}=10$ model pairs on each of the $60$ cases, yielding $600$ pairwise comparisons. Each comparison is judged by three of $12$ experts under blinded and randomized presentation. Human win rates use all judgments, with ties counting as half a win, and automatic scores are computed over the same cases. Human--automatic agreement includes only comparisons with a decisive A/B majority; majority-Tie and one-A/one-B/one-Tie outcomes remain in the win-rate calculation but are excluded from the agreement denominator. We report decisive-item coverage and 95\% case-clustered bootstrap confidence intervals in App.~Table~\ref{tab:human_alignment_item_level}. As a model-level summary, Spearman's~$\rho_s$ between human and automatic rankings ranges from $0.70$ to $1.00$ (Table~\ref{tab:human_alignment}). Full protocol details are provided in App.~\ref{app:human_alignment}.

\section{Conclusion}
We introduce \name, a comprehensive benchmark for keyframe-conditioned video generation. 
The benchmark comprises $386$ samples spanning four keyframe densities, two video structures, two prompt settings, two input formats, and three application domains, enabling systematic evaluation across diverse generation scenarios. 
The evaluation framework combines keyframe response metrics, which assess whether each input keyframe is reproduced at the intended temporal position and duration, with general quality metrics that evaluate overall video quality through a fully automated pipeline integrating vision-language model judges and specialized perception models.
Experimental results show that, although current models can faithfully reproduce individual keyframes under favorable conditions, they remain limited in following fine-grained instructions under dense keyframe constraints, generating coherent transitions between keyframes, decomposing storyboard-grid inputs (particularly in open-source models), and producing one-take videos. 
We hope \name serves as a standardized benchmark for diagnosing these limitations and facilitates future progress in keyframe-conditioned video generation.

{
    \small
    \bibliographystyle{ieeenat_fullname}
    \bibliography{main}
}

\clearpage
\setcounter{page}{1}
\maketitlesupplementary
\setcounter{section}{0}
\renewcommand{\thesection}{\Alph{section}}
 \UseRawInputEncoding

\section{Case Study}
\label{cases}
We present representative generated video results from both open-source and proprietary models, which can reflect their output quality and generation preferences. All examples should be read in row-major order.

\begin{figure*}[htbp]
     \centering
    \includegraphics[width=1\textwidth]{attachments/examples/closed/closed_exp_1.pdf}
    \caption{An example of a fable under minimal prompt. All models consistently preserve the visual style and closely follow the intended story progression specified by the input. Notably, Seedance 2.0 further enriches the sequence by introducing close-up shots of the lion’s attack from multiple viewpoints, resulting in a more dynamic and cinematic presentation.}
    \label{fig:close-1}
\end{figure*}

\subsection{Performance of Proprietary Models}
\label{closed-cases}
We present two examples for each prompt-control setting, with minimal-prompt cases shown in Fig.~\ref{fig:close-1} and Fig.~\ref{fig:close-2}, and segment-specific-prompt cases shown in Fig.~\ref{fig:close-3} and Fig.~\ref{fig:close-4}. Overall, proprietary models achieve strong qualitative performance, but they exhibit different generation preferences. Gemini-Omni-Flash does not always faithfully reproduce the input keyframes, yet it produces coherent and fluent videos with complete story progression and good instruction alignment. Seedance 2.0 shows the highest fidelity to the input keyframes, preserving visual details and temporal anchors more accurately. Kling-3.0-Omni provides a more balanced behavior between these two tendencies, maintaining relatively stable quality while preserving both keyframe correspondence and narrative coherence. Since Wan 2.7 only supports single-image storyboard-grid input, its output resolution and aspect ratio differ from those of the other models. Nevertheless, its results show that the model can generally interpret the storyboard grid correctly and organize the visual cues into a complete story.

\begin{figure*}[htbp]
     \centering
    \includegraphics[width=1\textwidth]{attachments/examples/closed/Closed_exp_3.pdf}
    \caption{An example of a cinematic story under minimal prompt. The outputs of all models generally reproduce the shot composition and camera staging specified by the input keyframes.}
    \label{fig:close-2}
\end{figure*}

\begin{figure*}[htbp]
     \centering
\includegraphics[width=1\textwidth]{attachments/examples/closed/Closed_exp_2.pdf}
    \caption{An example of a cinematic story under segment-specific prompt.}
    \label{fig:close-3}
\end{figure*}

\begin{figure*}[htbp]
    \centering
    \includegraphics[
        width=\textwidth,
        height=0.86\textheight,
        keepaspectratio
    ]{attachments/examples/closed/closed_exp_4.pdf}
    \caption{An example of a commercial under segment-specific prompt.}
    \label{fig:close-4}
\end{figure*}

\subsection{Performance of Open-Source Models}
\label{open-cases}
\begin{figure*}[htbp]
     \centering
    \includegraphics[width=1\textwidth]{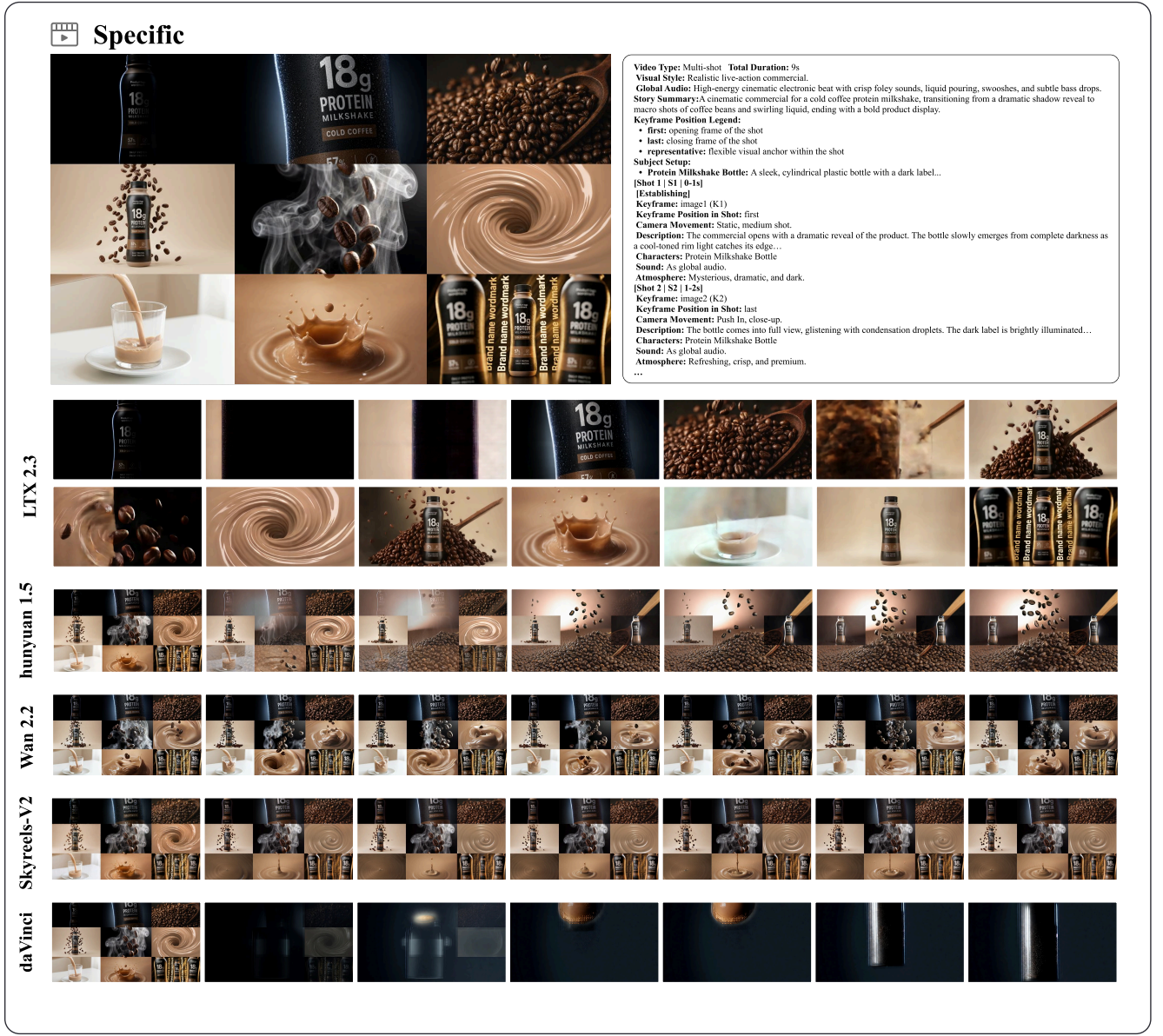}
    \caption{An example of a commercial advertisement under segment-specific prompt. This sample features rapid scene transitions with drastic differences between scenes. Although LTX 2.3 reproduces every single frame, its overall effect resembles PowerPoint slide transitions. The other models mostly produce static grid-like collage images.}
    \label{fig:open-1}
\end{figure*}

For each prompt-control setting, we also present two examples: minimal-prompt cases are shown in Fig.~\ref{fig:open-1} and Fig.~\ref{fig:open-2}, while segment-specific-prompt cases are shown in Fig.~\ref{fig:open-3} and Fig.~\ref{fig:open-4}. Among the open-source models, LTX 2.3 achieves the strongest overall performance, as it can recover and reproduce most input keyframes from the storyboard grid. However, its transitions between keyframes remain rigid, often appearing as hard cuts or slide-like changes rather than continuous video motion. In contrast, the other open-source models generally struggle to parse the storyboard grid into distinct keyframes, leading to static collage-like outputs instead of temporally coherent videos.

\begin{figure*}[htbp]
     \centering
    \includegraphics[width=1\textwidth]{attachments/examples/open/open_exp_2.pdf}
    \caption{An example under segment-specific prompt. In this sample, the visual differences between scenes corresponding to different shots are relatively small. Besides LTX 2.3, HunyuanVideo 1.5-I2V and daVinci-MagiHuman-1080p-I2V also produce some non-static results that go beyond simple storyboard-grid collages.}
    \label{fig:open-2}
\end{figure*}

\begin{figure*}[htbp]
     \centering
    \includegraphics[width=1\textwidth]{attachments/examples/open/open_exp_3.pdf}
    \caption{An example under minimal prompt. Some models produce abnormal and irrelevant content in this sample. Surprisingly, daVinci-MagiHuman-1080p-I2V correctly generates a single-shot video, although it does not fully respond to every input keyframe.}
    \label{fig:open-3}
\end{figure*}

\begin{figure*}[htbp]
     \centering
    \includegraphics[width=1\textwidth]{attachments/examples/open/open_exp_4.pdf}
    \caption{An example under minimal prompt. LTX 2.3 suffers from repeated texture artifacts, but it still achieves the best overall performance.}
    \label{fig:open-4}
\end{figure*}

\section{Calibration and Computation of Keyframe Similarity}
\label{app:kfs_details}
For each input keyframe $i$, we select the response frame with the
highest cosine similarity between $L_2$-normalized DINOv3 features:
\begin{equation}
\label{eq:kfs_correspondence}
\begin{aligned}
j_i^{\star}
&=
\operatorname*{arg\,max}_{j}
\hat{\mathbf{f}}_i^{\top}\hat{\mathbf{g}}_j, \\
c_i^{\mathrm{dino}}
&=
\hat{\mathbf{f}}_i^{\top}\hat{\mathbf{g}}_{j_i^{\star}},
\end{aligned}
\end{equation}
where $\hat{\mathbf{f}}_i$ and $\hat{\mathbf{g}}_j$ are the normalized
DINOv3 features of keyframe $i$ and response frame $j$, respectively.

We calibrate the semantic matching threshold on a held-out validation set of
50 samples spanning both prompt modes and the four short-video proprietary 
models. For each validation keyframe, we take the top-ranked
keyframe--response pair obtained by Eq.~\eqref{eq:kfs_correspondence} and label
it as matched or unmatched under a relaxed semantic-match criterion requiring
at least one salient subject, object, or event participant to be shared. This
yields 1,030 inspectable pairs; after excluding three uncertain cases, 1,027 binary labels are used for threshold selection.
We sweep the observed DINOv3 similarities and
select the threshold closest to the equal-error operating point:
\begin{equation}
\label{eq:kfs_threshold_calibration}
\gamma^{\star}
=
\operatorname*{arg\,min}_{t\in\mathcal{T}}
\left|
\operatorname{FPR}(t)-\operatorname{FNR}(t)
\right|,
\end{equation}
where $\mathcal{T}$ is the set of candidate thresholds. This procedure gives an equal-error operating point of
$\gamma=0.7837$, with an FPR of $5.3\%$ and an FNR of $5.4\%$.
For all benchmark runs, we use the rounded and slightly more conservative
operating threshold $\gamma^{\star}=0.80$; at this deployed threshold, the
validation FPR remains $5.3\%$ and the FNR is $6.4\%$. The threshold is fixed
before test evaluation.

The matched-keyframe set is
\begin{equation}
\label{eq:kfs_matched_set}
\mathcal{M}
=
\left\{
i
\,\middle|\,
c_i^{\mathrm{dino}}>\gamma^{\star}
\right\}.
\end{equation}

For each matched keyframe, we compute PSNR and SSIM against its selected
response frame. We empirically set the lower normalization anchors to
$\tau_p=12.0\,\mathrm{dB}$ and $\tau_s=0.65$. We use
$50\,\mathrm{dB}$ as the practical PSNR saturation value and $1.0$ as
the SSIM saturation value. The pixel-level similarity is
\begin{equation}
\label{eq:kfs_pixel_score}
\begin{aligned}
s_i^{\mathrm{pix}}
=
\max\Bigg\{
&
\operatorname{clip}_{[0,1]}
\left(
\frac{\mathrm{PSNR}_i-12}{50-12}
\right), \\
&
\operatorname{clip}_{[0,1]}
\left(
\frac{\mathrm{SSIM}_i-0.65}{1-0.65}
\right)
\Bigg\},
\end{aligned}
\end{equation}
where
\begin{equation}
\operatorname{clip}_{[0,1]}(x)
=
\min\{1,\max\{0,x\}\}.
\end{equation}
These parameters only define the scale of the continuous pixel score;
they do not determine whether a pair is semantically matched.

For each matched keyframe, we combine the pixel and semantic
similarities as
\begin{equation}
\label{eq:kfs_final}
\begin{aligned}
q_i
&=
0.4\,s_i^{\mathrm{pix}}
+
0.6\,c_i^{\mathrm{dino}},
\qquad i\in\mathcal{M}, \\
\mathrm{KFS}
&=
\frac{1}{N_{\mathrm{kf}}}
\sum_{i\in\mathcal{M}} q_i,
\end{aligned}
\end{equation}
where $N_{\mathrm{kf}}$ is the total number of input keyframes. Unmatched
keyframes are excluded from the sum but remain in the denominator, and
therefore contribute zero to the final score.

We further evaluate the sensitivity of KFS to the empirically selected pixel-score parameters. For this analysis, we use the subset with complete KFS
artifacts for all four short-video proprietary models, containing 52 minimal and 57 specific prompt-mode samples, 436 model outputs, and 2,892 input keyframes. We vary
$\tau_p\in\{10,12,14\}\,\mathrm{dB}$,
$\tau_s\in\{0.60,0.65,0.70\}$, and the PSNR saturation value in
$\{40,50,60\}\,\mathrm{dB}$, while keeping the semantic matches, DINOv3
threshold, SSIM saturation value, and fusion weights fixed. Across these
settings, the mean KFS changes by at most $\mathrm{0.0244}$, and the minimum
Spearman rank correlation with the default setting is $\mathrm{0.997}$,
showing that KFS is stable within the tested ranges.

\section{MLLM Judge Sampling and Stability}
\label{app:mllm_judge_stability}

\paragraph{Frame sampling.}
For all MLLM-judged metrics, each generated video is uniformly sampled at $8$ frames per second (FPS), with the sampled frames provided to the judge in chronological order. This rate retains short-lived motion changes and visual defects while keeping the input length manageable. To examine sensitivity to temporal sampling, we repeat the evaluation at $4$ and $2$ FPS on the same videos while keeping the judge model, prompt, and scoring procedure fixed. The resulting metric scores vary only slightly across sampling rates, without changing the overall conclusions. We therefore adopt $8$ FPS for all reported results.

\paragraph{Repeated-run stability.}
MLLM judgments may exhibit minor variation across repeated inference calls. We therefore run the MLLM-based evaluation five times at $8$ FPS on the $115$-sample common set used by the six-model short-video leaderboard, using identical video inputs, prompts, and judge configurations. Within each model and prompt mode, we first average each metric over valid samples and then compute Visual Quality as the mean of SVQ and DVQ, Spatiotemporal Coherence as the mean of AC, SOC, and PR, and Instruction Adherence as the mean of the applicable VMA and AMA scores; AMA is omitted under minimal prompts. Table~\ref{tab:mllm_judge_stability} reports the macro-average of each group across the six models and two prompt modes. The low standard deviations show that the aggregated scores are stable across independent judge invocations.

\begin{table*}[t]
\centering
\caption{Repeated-run stability of the MLLM-judged metric groups on the six-model, two-mode common set. The final column reports the mean and sample standard deviation across five independent runs.}
\label{tab:mllm_judge_stability}
\scriptsize
\setlength{\tabcolsep}{4.5pt}
\renewcommand{\arraystretch}{1.05}
\begin{tabular}{@{}lcccccc@{}}
\toprule
\textbf{Metric Group} & \textbf{Run 1} & \textbf{Run 2} & \textbf{Run 3} &
\textbf{Run 4} & \textbf{Run 5} & \textbf{Mean $\pm$ Std.} \\
\midrule
Video Quality & 0.6713 & 0.6728 & 0.6699 & 0.6717 & 0.6706 & $0.6713 \pm 0.0011$ \\
Spatiotemporal Coherence & 0.7448 & 0.7461 & 0.7439 & 0.7454 & 0.7438 & $0.7448 \pm 0.0010$ \\
Instruction Adherence & 0.7173 & 0.7192 & 0.7161 & 0.7180 & 0.7159 & $0.7173 \pm 0.0014$ \\
\bottomrule
\end{tabular}
\end{table*}

\section{Human Alignment Protocol}
\label{app:human_alignment}

\paragraph{Evaluation protocol.}
We use $60$ segment-specific test cases for which all five evaluated models have complete video outputs and automatic metric results. The final subset contains 10 one-take and 50 multi-shot cases, with 20, 20, 10, and 10 cases containing 3, 6, 9, and 12 keyframes, respectively. For each case, all $\binom{5}{2}=10$ unordered model pairs are constructed, yielding $600$ unique comparison items. Each item is rated independently by three annotators drawn from a pool of twelve domain experts. Annotators are shown the input keyframes, the segment-specific prompt, and two generated videos in randomized left--right order, with model identities and automatic scores hidden. For each dimension, they choose Video A, Video B, or Tie.

\paragraph{Human rubric.}
The human rubric follows the same observable scopes as the automatic metrics in Sec.~\ref{sec:benchmark:metric_design} and the grouped-judge rubrics in App.~\ref{sec:judge-prompt}. \textbf{Video Quality} combines the Static and Dynamic Visual Quality criteria, including clarity, rendering artifacts, color and style stability, discontinuity, repetition, implausible morphing, and flicker. \textbf{Spatiotemporal Coherence} combines Attribute Consistency, Spatial Orientation Consistency, and Physical Rationality. \textbf{Instruction Adherence} follows the Video and Audio Modality Adherence rubrics and evaluates only requirements explicitly specified by the segment-specific prompt; audio--visual timing contributes here only when requested by the prompt. \textbf{Audio-Visual Coordination} instead evaluates intrinsic semantic and temporal agreement between the generated audio and video, independent of the text prompt, matching the roles of ImageBind Similarity and JavisScore. A tie is used when neither video is consistently preferable under the current dimension.

\paragraph{Aggregation and alignment.}
After mapping the randomized left--right labels back to model identities, an individual preference contributes $1$ to the selected model and $0$ to its opponent, while a tie contributes $0.5$ to each. For each dimension, a model's human win rate is the mean over its four opponents, all $60$ cases, and three annotators, for $4\times60\times3=720$ judgments per model. The overall human score is the mean of the four dimension-level win rates. Nominal Krippendorff's $\alpha$ is computed separately within each unordered model pair, with Tie retained as a third category; the ten pair-specific values are summarized per dimension by their median and full range. Automatic scores are recomputed on the same $60$ cases using
\begin{equation}
\begin{aligned}
G_{\mathrm{VQ}} &= \tfrac{1}{2}(\mathrm{SVQ}+\mathrm{DVQ}), \\
G_{\mathrm{ST}} &= \tfrac{1}{3}(\mathrm{AC}+\mathrm{SOC}+\mathrm{PR}), \\
G_{\mathrm{IA}} &= \tfrac{1}{2}(\mathrm{VMA}+\mathrm{AMA}), \\
G_{\mathrm{AV}} &= \tfrac{1}{2}(\mathrm{IB}+\mathrm{JavisScore}).
\end{aligned}
\end{equation}
For each model and dimension, the model-level automatic score is the mean of the corresponding per-case score over the same $60$ cases; the overall automatic score is the mean of the four dimension-level scores. Human--automatic pairwise agreement is evaluated separately for each dimension. A comparison has a decisive human preference when at least two of its three annotators prefer the same model. Majority-Tie outcomes and one-A/one-B/one-Tie outcomes are excluded from the agreement denominator but retained in the human win-rate aggregation through their individual $0.5$ tie contributions. A decisive comparison agrees when the sign of its per-case automatic-score difference favors the same model; an exact automatic-score tie is counted as a disagreement with a decisive human preference. Decisive-item coverage is the number of decisive comparisons divided by all $600$ comparisons for that dimension. We use $10{,}000$ case-clustered bootstrap replicates, each resampling the $60$ cases with replacement while retaining all ten model pairs and three annotations belonging to each sampled case and then recomputing agreement and coverage. The 2.5th and 97.5th percentiles form the corresponding 95\% confidence intervals.

Spearman's $\rho_s$ is computed between the five model-level human win rates and automatic scores after converting each to ranks. All annotations, including ties, contribute to the human win rates used here; the decisive-majority filter applies only to pairwise agreement. Equal model-level scores receive average ranks. The two-sided exact $p$-value is obtained by enumerating all $5!=120$ permutations of the human model labels relative to the fixed automatic scores and counting permutations whose absolute correlation is at least the observed $|\rho_s|$.

\begin{table*}[t]
\centering
\caption{Item-level human--automatic agreement and annotation reliability. Agreement and decisive coverage are reported with case-clustered bootstrap 95\% confidence intervals. Pair-specific Krippendorff's $\alpha$ is summarized by the median and full range across the ten unordered model pairs.}
\label{tab:human_alignment_item_level}
\scriptsize
\setlength{\tabcolsep}{4pt}
\begin{tabular}{@{}lccc@{}}
\toprule
Dimension & Agreement [95\% CI] & Decisive coverage [95\% CI] & $\alpha$ median [range] \\
\midrule
Video Quality & 0.810 [0.770, 0.850] & 0.920 [0.890, 0.945] & 0.58 [0.44, 0.69] \\
Spatiotemporal Coherence & 0.777 [0.735, 0.820] & 0.880 [0.845, 0.910] & 0.52 [0.36, 0.64] \\
Instruction Adherence & 0.840 [0.805, 0.875] & 0.940 [0.915, 0.960] & 0.63 [0.51, 0.74] \\
AV Coordination & 0.729 [0.680, 0.780] & 0.830 [0.790, 0.865] & 0.45 [0.29, 0.58] \\
\bottomrule
\end{tabular}
\end{table*}

\section{Prompt Templates}
\label{sec:prompt}
This section presents the prompt templates used in \name, including the two video-generation prompt granularities, the generated-video segmentation contract, the per-metric checklist-generation prompt, and the grouped judge prompt with its scoring rubrics.

\subsection{Prompt Templates for Video Generation}
\label{sec:prompt-generation-video}
We present the two prompt granularities separately. Segment-specific prompts provide detailed constraints for each temporal segment. For multi-shot videos, the specifications are organized at the shot level, with each shot serving as the minimum unit. For single-shot videos, the prompt additionally includes a global camera-motion description, while the keyframes are treated as temporal anchor points and the content between each pair of adjacent keyframes is specified segment by segment.

\begin{promptlisting}{Segment-Specific Prompt for Single-Shot Video}
Video Type: One-Take
Total Duration: [total_duration]
Visual Style: [visual_style]
Global Audio: [global_audio]
Global Camera Motion: [overall_camera_motion]

Story Summary:
[brief_story_summary]

Input Keyframes:
The uploaded images define this one-take video as ordered keyframes: K1 → KN. Move through them in one continuous shot with smooth natural continuity while preserving each keyframe's appearance.

Keyframe Position Legend:
- first: this keyframe is the opening frame of the shot
- last: this keyframe is the closing frame of the shot
- representative: no strict timing constraints, use as a general visual anchor anywhere within the shot

Subject Setup:
- [Subject 1] ([role]): [appearance/identity description]
- ...

[Continuous Shot S1]

[t0 | Keyframe | image1 (K1)]
Frame: [visual description of keyframe K1]

[t0-t1 | Segment]
Camera Movement: [camera movement and framing]
Description: [motion/action transition from K1 to K2]
Characters: [subjects appearing in this segment]
Sound: [segment-specific audio when incremental sound requirements are specified, such as dialogue, internal monologue, or an alarm ringing]
Atmosphere: [emotional / visual tone]

[t1 | Keyframe | image2 (K2)]
Frame: [visual description of keyframe K2]
...

[tn | Keyframe | imagen (Kn)]
Frame: [visual description of final keyframe Kn]
\end{promptlisting}

\begin{promptlisting}{Segment-Specific Prompt for Multi-Shot Video}
Video Type: Multi-shot
Total Duration: [total_duration]
Visual Style: [visual_style]
Global Audio: [global_audio]

Story Summary:
[brief_story_summary]

Input Keyframes:
The uploaded images define the full video timeline as ordered keyframes: K1 → KN. Follow this order as one coherent story, with smooth natural continuity between adjacent keyframes while preserving each keyframe's appearance.


Keyframe Position Legend:
- first: this keyframe is the opening frame of the shot
- last: this keyframe is the closing frame of the shot
- representative: no strict timing constraints, use as a general visual anchor anywhere within the shot

Subject Setup:
- [Subject 1] ([role]): [appearance/identity description]
- ...

[Shot 1 | S1 | start_time-end_time]
[shot_function, such as 'Flash-Back'.]
Keyframe: [image_id] ([keyframe_id])
Keyframe Position in Shot: [first / last / representative]
Camera Movement: [camera movement and framing]
Description: [shot-level visual and action description]
Characters: [subjects appearing in this shot]
Sound:
[shot-specific audio when incremental sound requirements are specified, such as dialogue, internal monologue, or an alarm ringing]
Atmosphere: [emotional / visual tone]
...
\end{promptlisting}

Minimal prompts contain only the keyframe order, a brief story synopsis, the target video structure, and duration, encouraging the model to infer events and transitions mainly from the visual sequence.
\begin{promptlisting}{Minimal Prompt for Video Generation}
Video Type: Multi-shot/One-Take
Target duration: [total_duration]

Story Summary:
[brief_story_summary]

Input Keyframes:
The uploaded images define the full video timeline as ordered keyframes: K1 → KN. Follow this order as one coherent story, with smooth natural continuity between adjacent keyframes while preserving each keyframe's appearance.


Generation Goal:
Generate a coherent video that follows the keyframe order, filling in natural motion, state changes, and transitions between frames.
\end{promptlisting}

\subsection{Prompts for Video Segmentation}
\label{sec:segmentation}
Since the generated video structure may differ from the instructed structure, we use Gemini 3.1 Pro to segment each generated video and pre-assign keyframes to the resulting segments. This step lets prompt-specific requirements be matched to the actual generated-video segments rather than to the requested shot structure. The segmentation contract is:
\begin{promptlisting}{Contract for Video Segmentation}
Context:
The attached generated video is the only source of truth for its actual shot
structure. The benchmark spec describes the requested content and input
keyframes, but requested shot count, requested shot order, prompt wording, and
keyframe count are not evidence of cuts in the generated video.

Task:
Complete these two steps in order.

Step 1 - Observe and lock the actual video structure:
1. Watch the generated video without using the benchmark spec to decide where
   cuts should exist.
2. A shot is one continuous camera take. Start a new shot only at an actual
   edit boundary such as a hard cut, dissolve, wipe, or other visible transition
   to a different take.
3. Do not create a new shot for an action phase, subject state change, camera
   movement, reframing, occlusion, or prompt-requested event when the take
   remains continuous.
4. If the generated video has multiple actual shots, return one segment per
   actual shot in chronological order. If it is a one-take video, return
   exactly one segment covering the full video.

Step 2 - Assign keyframes after the shot boundaries are fixed:
1. Use the benchmark spec and visible content only to assign each input
   keyframe to every already-fixed actual shot time range that visibly contains
   or represents it.
2. Keyframe assignment must never add, remove, split, merge, shift, or reorder
   actual shot boundaries.
3. If no actual shot visibly responds to a keyframe, do not invent an
   assignment. List the keyframe in unassigned_keyframes instead.

Output contract:
1. Each segment is an actual generated-video shot, not an expected prompt shot
   and not a keyframe-specific semantic phase.
2. Use actual shot ids "S1", "S2", ... in chronological order. For an actual
   single-shot video, the only shot id is "S1", even when the benchmark spec
   requests multiple shots.
3. Use segment ids "SEG001", "SEG002", ... in the same chronological order.
4. Segment times are seconds in [0, duration_sec], sorted, contiguous,
   non-overlapping, and together cover the complete generated video.
5. Each input keyframe id may appear in zero, one, or multiple segments'
   assigned_keyframe_ids. If it appears in zero segments, it must appear exactly
   once in unassigned_keyframes.
6. A segment may contain zero, one, or multiple assigned keyframes.
7. unassigned_keyframes[*].reason must be exactly one of:
   "not_visible_in_generated_video" or "ambiguous". Use
   "not_visible_in_generated_video" only when the generated video has no
   visible response to that keyframe. Use "ambiguous" when the evidence is too
   weak or conflicting to assign a visible response to a segment.
8. Confidence reflects confidence in both the observed shot boundary and the
   keyframe assignment. Keep reason concise and evidence-based.

Do not:
- Copy expected shot ids from the benchmark spec unless they coincidentally
  match the actual generated-video shot order.
- Force the generated video to have the prompt-requested number of shots.
- Split a continuous single shot into action/state/keyframe phases.
- Merge actual shots merely because they depict the same subject or expected
  keyframe.

Examples:
- If the spec requests three shots but the generated video is one continuous
  take, return one full-duration segment with shot_id "S1" and assign all
  keyframes to that segment.
- If the spec requests a one-take video but the generated video visibly has
  two edited shots, return two segments with shot_ids "S1" and "S2", then
  assign each keyframe to the better-supported segment.

Benchmark spec JSON:
{spec_json}

Return strict JSON only. No markdown. No extra text.
Output shape:
{{
  "segments": [
    {{"segment_id": "SEG001", "shot_id": "S1", "assigned_keyframe_ids": ["K1"], "start_sec": 0.0, "end_sec": 2.4, "confidence": 0.83, "reason": "actual opening shot; K1 is most visible here"}}
  ],
  "unassigned_keyframes": [
    {{"keyframe_id": "K4", "reason": "not_visible_in_generated_video", "evidence": "no actual shot shows the K4 state"}}
  ]
}}
\end{promptlisting}

\subsection{Prompts for Checklist Generation}
\label{sec:checklists}
Checklist generation is performed independently for each metric. The shared template receives a filtered sample specification, a metric-specific construction procedure, and a metric-specific response shape. It asks the generator for sample-specific inspection lenses without exposing the generated video or asking it to assign scores.

\begin{promptlisting}{Per-Metric Checklist Generation Template}
You are the KeyFrame Bench checklist builder and rubric compiler for the [{metric_name}] metric.

Context:
You receive one benchmark sample: structured story, input keyframes, input shot descriptions, and filtered evaluation hints. A separate MLLM judge will later map your checklist to the generated video.

Task:
Build a sample-specific human-readable checklist for the requested metric. Your job is to define inspectable scoring lenses and item-level pass/partial/fail rules, not to score the generated video.

Requested metrics: {metric_name}
Backend provider: {provider}

Build the checklist:
- Write atomic items. Each item tests one observable claim.
- Derive each item from the structured sample spec, input keyframes, sample.shots, filtered evaluation_hints, and the metric procedure.
- Interpret filtered evaluation_hints narrowly: camera_motion_raw is the source camera text; camera_motion_term is the normalized camera term; camera_motion_class is an auxiliary camera/viewpoint class; scene_relation_from_previous says whether this shot starts the sample, continues the same scene, or remains in the same scene.
- Set reference_keyframe_ids to the input keyframes that anchor the claim. Use all relevant keyframes for cross-shot claims and [] for full-video, audio, or otherwise unanchored items.
- Return only dimension, reference_keyframe_ids, evidence_scope, human_question, and inspect for each checklist item. Runtime identifiers and metric metadata are added by the benchmark code.
- Use plain review language in human_question and inspect.
- Do not include item-level score anchors or pass/partial/fail rules; the judge applies the metric rubric when scoring.
- Prefer 4-8 checklist items when the metric procedure does not specify a count.
- For Attribute Consistency, fill subject_classification for character, animal, prop, and background routing.

Metric-specific rubric compilation procedure:
{procedures}

Evaluation spec JSON:
{spec_json}

Return strict JSON only. No markdown. No prose outside JSON.
Output shape:
{response_shape}
\end{promptlisting}

Checklist outputs follow schema v3. The model emits only the five semantic fields listed below; the benchmark code adds the schema version, metric name, and stable item identifier. Attribute Consistency additionally returns the top-level subject classification shown in its response shape.

\begin{promptlisting}{Checklist Response Field Contract}
schema_version: checklist_items_v3
required_item_fields:
  - dimension
  - reference_keyframe_ids
  - evidence_scope
  - human_question
  - inspect
\end{promptlisting}

The default response shape is used for all metrics except Attribute Consistency:

\begin{promptlisting}{Default Checklist Response Shape}
{
  "checklist_items": [
    {
      "dimension": "<metric-specific dimension>",
      "reference_keyframe_ids": ["K1"],
      "evidence_scope": "shot",
      "human_question": "One atomic, sample-specific review question.",
      "inspect": "Concrete visual or audio evidence to inspect."
    }
  ]
}
\end{promptlisting}

Attribute Consistency uses the same item contract and adds subject routing metadata:

\begin{promptlisting}{Attribute Consistency Checklist Response Shape}
{
  "subject_classification": {
    "subjects": [
      {
        "subject_id": "subject_1",
        "name": "young woman in red coat",
        "entity_type": "character",
        "evidence": "sample.shots and K1"
      }
    ],
    "has_character_subjects": true,
    "has_animal_subjects": false,
    "has_prop_subjects": false
  },
  "checklist_items": [
    {
      "dimension": "attribute_consistency.character",
      "reference_keyframe_ids": ["K1"],
      "evidence_scope": "shot",
      "human_question": "Across the segments where the woman appears, does her face identity remain recognizable from K1?",
      "inspect": "Compare facial geometry, eye shape, nose bridge, and skin tone against K1 whenever the face is visible."
    }
  ]
}
\end{promptlisting}

At runtime, the template's \texttt{procedures} placeholder is replaced by one of the following metric-specific fragments. Each fragment fixes the metric boundary, valid dimensions, evidence scope, item granularity, and question style.

\noindent\textbf{Static Visual Quality.} This procedure creates frame-level checks for clarity, rendering artifacts, lighting and composition, and style stability.
\begin{promptlisting}{Static Visual Quality Procedure}
Build checklist_items for static, frame-level visual quality.

Goal:
Create concrete visual inspection lenses for image-level fidelity in representative shots or segments. Prefer issues a reviewer can verify from still frames: blur, noise, local rendering artifacts, lighting/color, composition, and overall aesthetic quality.

Boundary:
This metric should only evaluate frame-level rendering and aesthetic quality. Use it for blur, low resolution, noise, banding, aliasing, edge bleeding, moire, local texture smearing, local texture corruption, exposure problems, lighting/color quality, composition quality, and style stability. Do not use it for body anatomy errors, extra/missing/fused parts, object structural collapse, contact/support/gravity/material behavior, prompt adherence, recurring identity drift, or spatial relationship consistency.

Cover these subdimensions when applicable. Use dimension prefix "static_quality.xxx":
- static_quality.clarity_fidelity: sharpness, resolution, readable details of main subjects, objects, text, faces, and important surfaces; absence of blur or local detail smearing.
- static_quality.artifacts: frame-level rendering artifacts such as noise, banding, aliasing, edge bleeding, moire, local texture smearing, local texture corruption, compression-like artifacts, or other image-level visual artifacts that do not change anatomy, object structure, or physical behavior.
- static_quality.color_lighting_aesthetics: visually coherent color, tone, exposure, contrast, lighting direction, shadows, and composition quality. Evaluate the aesthetic execution, not whether a prompt-specified color, lighting setup, or camera framing was followed.
- static_quality.style_consistency: stable visual rendering style across frames when the prompt/keyframes imply a consistent style. Do not use this for identity, clothing, prop, or background attribute drift.

Item construction:
- Prefer 3-6 checklist_items.
- Use "shot" or "segment" scope for localized frame quality checks.
- Use "full_video" only for style consistency or quality issues visible throughout the output.
- Each item tests one subdimension and one observable visual criterion.

Good human_question pattern:
"During S1, is the main subject rendered with sharp, readable detail without obvious blur or local texture smearing?"
\end{promptlisting}

\noindent\textbf{Dynamic Visual Quality.} This procedure assigns each temporal artifact to one owner dimension, separating discontinuities, stalled playback, implausible morphing, flicker, and boundary glitches.
\begin{promptlisting}{Dynamic Visual Quality Procedure}
Build checklist_items for dynamic temporal rendering quality.

Goal:
Create review lenses for whether the generated video has clean temporal playback, independent of whether it followed the prompt, whether the motion is physically natural, or whether subject spatial relationships remain consistent.

Cover these subdimensions when applicable. Use dimension prefix "dynamic_quality.xxx". Each observable artifact must be assigned to exactly one subdimension; follow the ownership rules below so the same artifact is never penalized twice:
- dynamic_quality.frame_discontinuity: abrupt spatial jumps between adjacent frames — skipped-frame jumps, teleportation-like position changes, and slideshow-style hard replacements where content is swapped without intermediate frames. Owns the jump/replacement moment itself, including any position change caused by missing or broken intermediate frames (even when an object appears to melt away and reappear elsewhere). Does not own the frozen or repeated interval that may precede a jump — that belongs to stutter_repetition.
- dynamic_quality.stutter_repetition: intervals where motion stalls in time — unintended stutter, frozen intervals, held/static frames, and repeated frames that substitute for expected continuous motion. Owns only the stagnant interval; if a freeze ends with an abrupt jump, the jump itself is judged under frame_discontinuity. Hard slideshow-style replacements also belong to frame_discontinuity, not here.
- dynamic_quality.implausible_morphing: unintended in-place frame-to-frame melting, dissolving, ghost blending, or morphing where objects, hands, text, packages, surfaces, or subject parts smear into a different state instead of moving continuously. Owns only shape/appearance corruption at roughly the same location; sudden position changes — even ones that pass through a melted or blended intermediate look — belong to frame_discontinuity.
- dynamic_quality.flicker_instability: sustained or recurring flicker within a shot — strobe, rapid brightness popping, color popping, or repeated frame-to-frame visual flashing that disrupts interpretation. Does not own single flash frames at segment or shot boundaries — those are boundary glitches and belong to a cross_shot item.

Item construction:
- Prefer 3-5 checklist_items.
- Use "full_video" for recurring temporal artifacts across the output.
- Use "cross_shot" only for unintended temporal glitches localized at segment or shot boundaries, such as single flash frames, skipped-frame jumps at the cut, or repeated frames at the cut. Boundary glitches are owned by the cross_shot item and must not also be counted by a full_video item.
- Use a specific segment only when a visible temporal artifact is localized.
- Each item tests exactly one subdimension: one temporal artifact type and one observable criterion. Do not combine multiple artifact types in a single human_question.

Good human_question patterns (one artifact type per item):
"Across the full video, are the main subjects free from skipped-frame jumps or teleportation-like position changes between adjacent frames?"
"Across the full video, does playback remain free of unintended frozen intervals or repeated frames that substitute for continuous motion?"
"Across the full video, do objects and subject parts keep their form without in-place melting, dissolving, or ghost blending between frames?"
"Within each shot, is the video free of flicker, strobing, or rapid brightness/color popping?"
"At each segment boundary, is the transition free of flash frames, repeated frames, or skipped-frame jumps introduced by the cut?"
\end{promptlisting}

\noindent\textbf{Attribute Consistency.} This procedure classifies recurring characters, animals, and props, then creates identity or attribute persistence checks anchored to all relevant keyframes.
\begin{promptlisting}{Attribute Consistency Procedure}
First classify expected subjects from story_summary, global_subjects, shots, and keyframes:
- character: visible human, humanoid, or named person.
- animal: non-human living creature.
- prop: object, vehicle, tool, garment, food, or product.

Classify by visible entity type, not by section labels such as "Characters". Products, tools, vehicles, garments, food, UI panels, watch faces, product fronts, and object bodies are props unless a visible human/humanoid/animal is present. Set has_character_subjects true only when at least one subject is classified as character.

Fill subject_classification with subjects, entity_type, and evidence.

Build checklist_items for identity and attribute persistence.

Goal:
Select the most important recurring entities and attributes that should remain recognizable across the relevant video span. Accept changes explicitly requested by the prompt, then check whether the new state persists.

Use these dimensions, matching the entity_type of the checked subject. Use dimension prefix "attribute_consistency.xxx":
- attribute_consistency.character: face structure, age, hairstyle, clothing, accessories, body silhouette.
- attribute_consistency.animal: species, body shape, fur/skin color, markings.
- attribute_consistency.prop: shape, color, size, material, logo, distinctive details.

Item construction:
- Prefer 4-8 checklist_items for samples with multiple recurring subjects.
- Prioritize entities visible in keyframes or central to the prompt.
- Each item names one entity and one attribute or identity cue.
- Include every input keyframe that visibly anchors the attribute in reference_keyframe_ids.

Good human_question pattern:
"Across the segments where the woman appears, does her red coat remain recognizable in color and silhouette unless the prompt requests a change?"
\end{promptlisting}

\noindent\textbf{Spatial Orientation Consistency.} This procedure checks visible relative positions and scale proportions while allowing changes explained by subject motion, viewpoint changes, or scene transitions.
\begin{promptlisting}{Spatial Orientation Consistency Procedure}
Build checklist_items for internal spatial-relation consistency.

Goal:
Check whether subject/object and subject/environment spatial relations remain self-consistent given the visible camera viewpoint and visible actions.

Boundaries:
- Do not create spatial-relation items whose main question is whether the camera followed the prompt.
- Use the visible shot descriptions, keyframes, generated video, and filtered evaluation_hints to frame the spatial expectation. scene_relation_from_previous can identify starts/new scenes/same-scene continuity, and camera_motion_class can explain apparent layout changes caused by viewpoint motion; distinguish viewpoint motion from true subject/object drift.
- Do not create items about motion smoothness, jitter, flicker, or repeated frames unless they directly obscure a spatial relation.
- Do not create items whose main failure is contact, grip, clipping, collision, support, or material behavior.

Cover these subdimensions when visually grounded. Use dimension prefix "spatial_orientation.xxx":
1. spatial_orientation.relative_position_stability:
   - left/right, front/behind, near/far, facing direction, containment, and ordering between named subjects remain stable when no visible action explains a change.
   - Judging rule: a relation change is acceptable when visible subject movement, camera viewpoint change, or an explicit scene transition explains it; treat only unexplained changes as failures.
2. spatial_orientation.scale_proportion_rationality:
   - Proportional size between subjects, objects, and environment landmarks remains plausible across camera scales and angles.

Item construction:
- Prefer 2-5 checklist_items.
- Name the two subjects or subject/environment pair being checked.
- Use only relations with visible evidence in keyframes, generated frames, or the attached video. Do not invent abstract geometry not visible in media.
- Use "cross_shot" for relation/proportion checks spanning segments.
- Use "full_video" only when a spatial relation must remain stable throughout.
- Use dimensions exactly from the subdimension names above.

Good human_question pattern:
"Between S1 and S2, does the character remain on the same side of the table unless a visible movement explains the change?"
\end{promptlisting}

\noindent\textbf{Physical Rationality.} This procedure defines broad checks for interaction, anatomy, non-biological structure, important text readability, and scene-level common sense without double-counting one failure.
\begin{promptlisting}{Physical Rationality Procedure}
Build checklist_items for general physical plausibility across the generated video.

Goal:
Create focused judging lenses for visible physical and commonsense plausibility failures across the generated video, including issues that were not predictable from the prompt or keyframes. Each checklist item should test one clearly bounded failure type so different dimensions do not duplicate the same issue.

Prefer 3-5 checklist_items. Use evidence_scope "full_video" for the default case, and use "cross_shot" only when the physical issue is inherently about continuity between visual segments.

Use only the exact dimensions below. Use dimension prefix "physical_rationality.xxx":
- physical_rationality.interaction_plausibility: people, objects, and the environment maintain plausible contact, depth, occlusion, grasping, collision, and support. Use this for clipping or pass-through only when separate entities incorrectly occupy the same space during interaction.
- physical_rationality.biological_anatomy_violation: humans, animals, and other living bodies keep a natural and complete body structure. Use this for extra, missing, fused, detached, or impossibly bent body parts; do not use it for object contact failures unless the body itself becomes structurally impossible.
- physical_rationality.non_biological_structural_loss: objects, vehicles, buildings, tools, and other non-living things keep a stable and recognizable structure. Use this for disappearing parts, fused components, impossible holes, melted shapes, broken containment, or structural collapse.
- physical_rationality.scene_level_common_sense: character behavior, event logic, and basic world axioms make sense to a normal viewer. Use this for logically impossible or context-breaking behavior, such as an audience facing away from the stage while watching a performance, an agent signaling escape but running toward visible danger without explanation, or the sun reversing an established east-west path.
- physical_rationality.text_readability_common_sense: visible text-like content that should be readable in context, such as signs, labels, captions, books, posters, or screens, forms coherent stable text rather than nonsensical glyphs, scrambled letters, melted strokes, or frame-to-frame text corruption.

Boundary rules:
- Use scene_level_common_sense for behavior, event logic, and world axioms, not for failures already covered by contact, anatomy, object structure, or text readability.
- Use text_readability_common_sense for commonsense text legibility only when text is visible and contextually important. Do not create this item for text that is merely small, out of focus, low-resolution, compressed, or over/under exposed; use it when the text itself becomes nonsensical, scrambled, melted, or unstable across frames.
- If a failure could fit multiple dimensions, choose physical interaction, biological anatomy, non-biological structure, or text readability before scene-level common sense.

Human_question guidance:
Questions should judge a broad property across the whole video or a major segment. Do not limit a question to one expected action, object, or moment unless that issue is inherently cross-shot.

Good human_question patterns:
"Across the video, do people, objects, and the environment maintain plausible contact, depth, and occlusion without obvious clipping or pass-through?"
"Across the video, do living bodies remain anatomically complete and natural without extra, missing, fused, detached, or impossibly bent body parts?"
"Across the video, do non-living objects and environments preserve a stable and recognizable structure without unexplained deformation or missing parts?"
"Across the video, does visible important text, signage, labels, or screen content remain readable as coherent text rather than becoming scrambled or nonsensical?"
"Across the video, do character behavior, reactions, event logic, and basic world behavior remain understandable and consistent with common sense?"
\end{promptlisting}

\noindent\textbf{Video Modality Adherence.} This procedure converts explicit video instructions into checks for camera behavior, shot structure, narrative rhythm or focus, and required segment content.

\begin{promptlisting}{Video Modality Adherence Procedure}
Build checklist_items for whether the generated video follows the prompt/spec.

Goal:
Evaluate requested video content and structure: camera instructions, shot count/structure, narrative rhythm, story focus, and required subject/prop/scene presence. Do not evaluate internal visual quality, spatial consistency, or physical plausibility.

Cover these dimensions when applicable. Use dimension prefix "video_adherence.xxx":
1. camera_parameter (dimension: "video_adherence.camera_parameter"):
   - camera motion, scale, angle, or framing matches the prompt/spec.
2. shot_count_structure (dimension: "video_adherence.shot_count_structure"):
   - generated shot count or continuous-take structure matches the prompt/spec.
   - for reviewer-facing human_question and inspect text, call a single-shot structure "one-take (a single continuous shot with no cuts)". Do not use the internal label "single-shot" by itself.
   - when shot count is the issue, describe multi-shot as "multiple distinct shots separated by cuts or transitions".
3. narrative_logic (dimension: "video_adherence.narrative_rhythm" or "video_adherence.story_focus"):
   - transitions and duration distribution support the intended story focus.
4. segment_presence (dimension: "video_adherence.segment_presence"):
   - required characters, props, actions, and scenes appear in the intended segments.

Item construction:
- Prefer 3-7 checklist_items.
- Each item tests one requested prompt/spec requirement.
- Use "shot" or "segment" for localized requirements.
- Use "cross_shot" for transition/rhythm checks.
- Use "full_video" for overall story-focus or structure checks.

Good human_question pattern:
"Across the full video, does the generated sequence preserve the prompt's intended story focus rather than spending most time on unrelated actions?"
\end{promptlisting}

\noindent\textbf{Audio Modality Adherence.} This procedure creates audio-only checks for requested sounds, dialogue, music, ambience, style, or emotion; it marks the metric not applicable when no audio requirement exists.
\begin{promptlisting}{Audio Modality Adherence Procedure}
Build checklist_items for whether generated audio follows the prompt/spec.

Goal:
Evaluate audio requirements only when the prompt/spec includes sound, music, dialogue, or ambience expectations. Do not evaluate audio-visual synchronization.

Cover these dimensions when applicable. Use dimension prefix "audio_adherence.xxx":
1. semantic_alignment (dimension: "audio_adherence.semantic_alignment"):
   - requested sounds, music, dialogue, or ambience are present and match what the prompt/spec asks for.
2. style_emotion (dimension: "audio_adherence.style_emotion"):
   - overall audio style, mood, genre, or emotional tone matches the prompt/spec's intended atmosphere.

Item construction:
- Use evidence_scope "audio_track".
- Prefer 1-3 checklist_items when audio requirements exist.
- If no audio requirement exists, create one item stating that no audio-specific requirement is present and set evidence_scope to "not_applicable".
\end{promptlisting}

For minimal-prompt runs, the builder substitutes the following specialized fragments for the three metrics whose standard procedures depend on explicit shot-level instructions. These variants restrict checklist construction to subjects, relations, structure, and narrative expectations grounded in the keyframes and story summary.

\begin{promptlisting}{Video Modality Adherence Minimal-Prompt Procedure}
Build checklist_items for whether the generated video satisfies the minimal prompt's visible subjects, story summary, and broad video structure.

Goal:
Use keyframes, story_summary, and subject_classification from Attribute Consistency to evaluate instruction adherence without inventing details absent from the minimal prompt.

Cover these dimensions. Use dimension prefix "video_adherence.xxx":
1. segment_presence (dimension: "video_adherence.segment_presence"):
   - all core subjects listed in subject_classification.subjects appear in the generated video unless the story_summary clearly makes one background-only.
   - required central characters, props, animals, vehicles, or scene elements are visually present enough for a judge to recognize them.
2. shot_count_structure (dimension: "video_adherence.shot_count_structure"):
   - multi-shot samples contain multiple distinct shots separated by cuts or transitions.
   - one-take samples remain a single continuous shot with no cuts unless the keyframes imply a transition.
   - for reviewer-facing human_question and inspect text, use "one-take (a single continuous shot with no cuts)" instead of exposing the internal label "single-shot" by itself.
3. narrative_logic (dimension: "video_adherence.narrative_rhythm" or "video_adherence.story_focus"):
   - visual differences between adjacent keyframes inform expected rhythm.
   - the generated video represents the main events implied by keyframes and story_summary.

Item construction:
- Prefer 3-5 checklist_items.
- Each item tests one visible subject, structural, or narrative expectation.
- Use evidence_scope "full_video" for subject presence and story-focus checks.
\end{promptlisting}

\begin{promptlisting}{Attribute Consistency Minimal-Prompt Procedure}
Extract identifiable subjects from the input keyframe images and evaluate appearance consistency throughout the generated video.

Step 1: Classify visible subjects from keyframe images:
- character: visible human or humanoid.
- animal: visible non-human living creature.
- prop: distinctive object, vehicle, or item.

Ignore prompt section labels such as "Characters" when choosing entity_type. Product faces, watch faces, front panels, interfaces, surfaces, and object bodies remain props/background unless a visible human/humanoid/animal appears. Set has_character_subjects true only when at least one subject is classified as character.

Step 2: Fill subject_classification with identified subjects.

Step 3: Build checklist_items.

Item construction:
- Prioritize visible subjects that recur across keyframes or appear central to the story_summary.
- Check one entity and one identity/attribute cue per item.
- Use dimension "attribute_consistency.character", "attribute_consistency.animal", or "attribute_consistency.prop", matching the entity_type of the checked subject.
- Include every input keyframe that visibly anchors the attribute in reference_keyframe_ids.
\end{promptlisting}

\begin{promptlisting}{Spatial Orientation Consistency Minimal-Prompt Procedure}
Build checklist_items for internal spatial-relation consistency inferred from input keyframe images.

Goal:
When two subjects or a subject/environment relation is visible in keyframes, check that the generated video preserves the relation unless a visible action explains a change.

Cover these aspects. Use dimension prefix "spatial_orientation.xxx":
1. relative_position_stability (dimension: "spatial_orientation.relative_position_stability"):
   - left/right, front/behind, near/far, facing, and containment relations.
   - Judging rule: a relation change is acceptable when visible subject movement, camera viewpoint change, or a scene transition explains it; treat only unexplained changes as failures.
2. scale_proportion_rationality (dimension: "spatial_orientation.scale_proportion_rationality"):
   - proportional size between subjects and environment remains plausible across visual segments.

Item construction:
- Use only relations with clear visual evidence in at least one keyframe.
- Name the subjects and relation being checked.
- Use "cross_shot" when comparing visual segments.
- Use dimensions exactly as specified above.
\end{promptlisting}

\subsection{Prompts for Grouped Judge}
\label{sec:judge-prompt}
The evaluation stage supplies the grouped judge with the generated video, input keyframes, the filtered evaluation context, and the precomputed checklists. The template resolves referenced keyframes through judge grounding, applies the matching rubric to every checklist item, and requires timestamped observable evidence. Related metrics are evaluated together, but each metric retains its own checklist results and score.

\begin{promptlisting}{Grouped Judge Template}
You are the KeyFrame Bench evidence-anchored MLLM judge for the [{group_name}] metric group.

Context:
You receive the generated video, input keyframes, prompt/spec JSON, and a prebuilt checklist for a benchmark metric group. The checklist contains the required scoring lenses for each requested metric.

Task:
For every requested metric, score its checklist items from observable evidence, then write the metric-level score, reason, and evidence summary from the full evidence you reviewed.

Checklist and metric rules:
- Treat checklist_items as the required scoring lenses for each requested metric.
- Each checklist item is defined by its dimension, human_question, inspect, evidence_scope, and reference_keyframe_ids fields. Resolve referenced input keyframes through judge_grounding, then score with the matching metric rubric below.
- For Physical Rationality, use the checklist as broad lenses for the full video. When a clearly visible physical impossibility appears outside the wording of a single item, include it in the nearest relevant check_result evidence and in the metric-level reason.
- Score every checklist item unless visual/audio evidence is genuinely insufficient; then set that item score to null and explain the insufficiency.
- Metric-level score should be the arithmetic mean of valid check_results for that metric. The parser will recompute this, so consistency matters.

Evidence contract:
- Score each checklist item from observable evidence, not overall impression.
- For every check_result, include at least one evidence object with timestamp_sec or time_range_sec when the video/audio is the evidence.
- Good evidence is a concrete observation: who/what, where in time, and how it supports pass/partial/fail.
- Generic evidence such as "looks good", "mostly follows", or "there are artifacts" is not sufficient without timestamp and observable detail.
- If the full video is attached, prefer time_range_sec over frame ids.
- If Checklist JSON contains sensor_context, use it only as auxiliary context. For MonST3R camera context, read plain_language_summary, judge_guidance, score_conversion_rule, confidence_policy, and coordinate_caveat first. Then use estimated_camera_motion_label, confidence, and smoothness_description only as supporting camera-motion hints to compare against the visible video and any prompt camera requirement. MonST3R axes are internal, so verify real screen direction from the video; never copy a MonST3R value or smoothness label into a checklist score, and never treat it as an object-relation, mask, bbox, depth, or pose graph.

Scoring procedure:
- Use normalized score in [0, 1].
- Apply the matching metric rubric below to the evidence for each checklist item. Return a numeric score that reflects the observed severity for that item's dimension.
{rubric_section}
Backend provider: {provider}

Evaluation spec JSON:
{spec_json}

Checklist JSON:
{checklist_json}

Return strict JSON only. No markdown. No prose outside JSON.
Output shape. Repeat one metric object for each metric represented in Checklist JSON, and include one check_results entry for each checklist item:
{response_shape}
\end{promptlisting}

The judge returns one object for each metric represented in the checklist. Each item result copies the stable checklist identifier, records a normalized score and timestamped evidence, and may cite auxiliary tool evidence. The metric-level score is the arithmetic mean of valid item scores and is recomputed by the response parser.

\begin{promptlisting}{Judge Response Shape}
{
  "metrics": {
    "<metric name from Checklist JSON>": {
      "score": 0.0,
      "reason": "summary of check_results for this metric",
      "evidence": ["short metric-level evidence summary"],
      "used_tool_evidence_ids": [],
      "check_results": [
        {
          "checklist_id": "<copy checklist item id exactly>",
          "score": 0.75,
          "evidence": [
            {
              "time_range_sec": [0.0, 1.2],
              "observation": "concrete observed evidence with timestamp",
              "supports": "pass"
            }
          ],
          "reason": "rubric-based reason for this checklist item",
          "used_tool_evidence_ids": []
        }
      ]
    }
  }
}
\end{promptlisting}

The template's \texttt{rubric\_section} placeholder is filled with the following rubric for every metric in the group. The rubrics define ownership boundaries and normalized score anchors; they do not introduce additional checklist fields.

\begin{promptlisting}{Static Visual Quality Rubric}
Static Visual Quality rubric:
Judge frame-level image quality only, using the checklist item's dimension and scope.

Dimension cues:
- static_quality.clarity_fidelity: sharpness, resolution, readable details, face/text/object legibility, and absence of blur or local smearing.
- static_quality.artifacts: noise, banding, aliasing, edge bleeding, moire, compression-like damage, or local texture corruption.
- static_quality.color_lighting_aesthetics: exposure, contrast, tone, lighting direction, shadows, and composition quality.
- static_quality.style_consistency: stable rendering style across the scoped frames when a consistent style is expected.

Score anchors:
  score 1.0: The scoped frames are clean for this dimension; important details are readable and defects are absent or trivial.
  score 0.75: Mostly clean; only minor local defects appear, and they do not reduce trust in the visual evidence.
  score 0.5: Noticeable degradation is present, but the main content remains readable and usable.
  score 0.25: Severe or repeated defects make important subjects, objects, text, surfaces, or style cues hard to trust.
  score 0.0: The scoped evidence is unusable because of heavy blur, distortion, artifacts, exposure/color failure, or style collapse.
\end{promptlisting}

\begin{promptlisting}{Dynamic Visual Quality Rubric}
Dynamic Visual Quality rubric:
Judge temporal rendering quality only, using the checklist item's dimension and scope.

Dimension cues:
- dynamic_quality.frame_discontinuity: skipped-frame jumps, abrupt temporal pops, or teleportation-like changes caused by broken interpolation.
- dynamic_quality.stutter_repetition: unintended stutter, frozen intervals, held/static frames, repeated frames, slideshow-like transitions, or slide-deck-style hard replacements that interrupt playback or substitute for expected continuous motion.
- dynamic_quality.implausible_morphing: unintended frame-to-frame melting, dissolving, ghost blending, or morphing of objects, hands, text, packages, surfaces, or subject parts instead of continuous motion.
- dynamic_quality.flicker_instability: brightness flicker, color popping, strobe-like flashing, or frame-to-frame instability.

Score anchors:
  score 1.0: Playback is temporally clean for this dimension; motion or transitions stay continuous without meaningful artifacts.
  score 0.75: Mostly clean; a brief or low-impact artifact is visible but does not disrupt interpretation.
  score 0.5: Noticeable temporal artifacts occur, but the action or transition remains understandable.
  score 0.25: Frequent or severe artifacts make motion, transitions, or state changes hard to follow.
  score 0.0: Discontinuity, repetition, freezing, implausible morphing, flicker, or frame popping dominates the scoped evidence.
\end{promptlisting}

\begin{promptlisting}{Attribute Consistency Rubric}
Attribute Consistency rubric:
Judge whether the named entity and named identity/attribute cue remain recognizable across the checklist item's scope.

Dimension cues:
- character: face structure, age, hairstyle, clothing, accessories, and body silhouette.
- animal: species, body shape, fur/skin color, markings, and distinctive features.
- prop: shape, color, size, material, logo, details, and object identity.

Score anchors:
  score 1.0: The entity and requested cue stay stable and match the relevant prompt/keyframe reference, except for prompt-requested changes.
  score 0.75: Mostly stable; minor drift or brief ambiguity appears, but the same entity and cue remain clear.
  score 0.5: Partly stable; the entity or cue is still identifiable, but drift, substitution, or intermittent loss is obvious.
  score 0.25: Weakly stable; identity or attribute cues are often unstable or only match under generous interpretation.
  score 0.0: The entity is missing, replaced, contradicted by the reference, or the named attribute is absent or persistently wrong.
\end{promptlisting}

\begin{promptlisting}{Spatial Orientation Consistency Rubric}
Spatial Orientation Consistency rubric:
Judge whether the named spatial relation or proportion stays internally consistent after accounting for visible movement, scene changes, and camera viewpoint.

Dimension cues:
- relative_position_stability: left/right, front/behind, near/far, facing, containment, and ordering between named subjects or subject/environment pairs.
- scale_proportion_rationality: plausible relative size between subjects, objects, and environment landmarks across camera scales and angles.
- viewpoint_explained_relation_change: apparent relation changes are explainable by camera movement, viewpoint, visible subject movement, or an explicit scene transition.
- MonST3R sensor_context, when present, is only auxiliary camera-motion context; it does not provide object masks, boxes, depth, or pose evidence.

Score anchors:
  score 1.0: The relation, ordering, facing, containment, or proportion is stable where expected, and apparent changes are clearly explained.
  score 0.75: Mostly consistent; one brief ambiguity, occlusion, or camera-motion uncertainty does not break the relation.
  score 0.5: Partly consistent; the relation or proportion becomes unstable or ambiguous at times, but the layout remains interpretable.
  score 0.25: Weak spatial logic; relations often flip, collapse, reset, or show proportions that viewpoint/movement cannot explain.
  score 0.0: The named relation is contradicted, the layout resets without explanation, or scale/proportion becomes impossible.
\end{promptlisting}

\begin{promptlisting}{Physical Rationality Rubric}
Physical Rationality rubric:
Judge visible physical and common-sense plausibility. Use the checklist dimension to assign the main failure type; do not double-count one problem under several dimensions.

Dimension cues:
- physical_interaction_plausibility: contact, depth, occlusion, grasping, collision, and support between people, objects, and environment.
- biological_anatomy_violation: complete, natural body structure for humans, animals, and other living bodies.
- non_biological_structural_loss: stable structure for objects, vehicles, buildings, tools, and environments.
- gravity_support_materials: believable weight, gravity, balance, support, cloth, liquid, fire, smoke, deformation, and material behavior.
- text_readability_common_sense: visible important signs, labels, captions, books, posters, or screen text remain readable as coherent stable text, not scrambled glyphs, melted strokes, or nonsensical frame-to-frame corruption.
- scene_level_common_sense: behavior, reactions, event logic, and basic world axioms when no narrower physical dimension fits better.

Score anchors:
  score 1.0: No meaningful physical or common-sense break is visible for this dimension in the required scope.
  score 0.75: Mostly plausible; one or two minor local issues do not change interpretation.
  score 0.5: Noticeable physical errors occur, but the intended interaction, structure, material behavior, text readability, or event logic remains understandable.
  score 0.25: Repeated, severe, or distracting errors make the scoped physical behavior hard to accept.
  score 0.0: Impossible or contradictory physical behavior dominates, or the scene logic breaks in an interpretation-changing way.
\end{promptlisting}

\begin{promptlisting}{Video Modality Adherence Rubric}
Video Modality Adherence rubric:
Judge whether the generated video follows the prompt/spec requirement named by the checklist item. For minimal prompts, judge only structure or narrative expectations inferable from keyframes/story_summary.

Dimension cues:
- camera_parameter: requested camera motion, scale, angle, framing, or continuous-take behavior. MonST3R sensor_context, when present, is auxiliary camera-motion context; the visible video remains primary.
- shot_count_structure: requested shot count, multi-shot separation, one-take structure (a single continuous shot with no cuts), or scene-cut behavior. "single-shot" is an internal structure label; reviewer-facing checklist text should use the clearer term "one-take (a single continuous shot with no cuts)".
- narrative_logic.rhythm: duration distribution, transition pacing, and whether key events receive appropriate visual time.
- narrative_logic.story_focus: whether the video stays focused on the intended story rather than unrelated actions or scenes.
- segment_presence: required characters, props, actions, and scenes appear in the intended shot/segment.

Score anchors:
  score 1.0: The requirement is clearly present in the intended scope with correct timing, placement, structure, and semantic content.
  score 0.75: Mostly adherent; only minor timing, localization, framing, intensity, or completeness issues are visible.
  score 0.5: Partly adherent; the core intent appears but is incomplete, weakly localized, compressed, overextended, or missing an important detail.
  score 0.25: Weakly adherent; the requirement appears briefly, ambiguously, in the wrong segment, or only by generous interpretation.
  score 0.0: The requirement is missing, contradicted, placed in the wrong structure/segment, or replaced by unrelated content.
\end{promptlisting}

\begin{promptlisting}{Audio Modality Adherence Rubric}
Audio Modality Adherence rubric:
Judge whether generated audio follows the prompt/spec requirement named by the checklist item. If the item is explicitly not_applicable because no audio requirement exists, follow that item instead of inventing an audio expectation.

Dimension cues:
- audio_semantic_alignment: requested sounds, music, dialogue, ambience, or sonic events are audible and match the described visual events.
- audio_style_emotion: mood, genre, intensity, style, and emotional tone fit the requested scene.
- audio_visual_sync: audible events align with visible causes, impacts, speech, footsteps, actions, or timing cues when sync is implied.

Score anchors:
  score 1.0: The audio requirement is clearly satisfied, audible, semantically matched, stylistically appropriate, and synchronized when needed.
  score 0.75: Mostly adherent; only minor timing, style, intensity, clarity, or localization mismatch is present.
  score 0.5: Partly adherent; audio supports the scene but misses one important requested property, or sync/style/semantics are only partial.
  score 0.25: Weakly adherent; audio is generic, ambiguous, poorly synchronized, faint, or only indirectly related to the requested sound.
  score 0.0: Audio is missing, inaudible, contradicted by the requirement, unusable, or dominated by unrelated sound.
\end{promptlisting}

\end{document}